# A Survey on Offline Reinforcement Learning: Taxonomy, Review, and Open Problems

Rafael Figueiredo Prudencio, Marcos R. O. A. Maximo, and Esther Luna Colombini, *Member, IEEE*

*Abstract*— With the widespread adoption of deep learning, reinforcement learning (RL) has experienced a dramatic increase in popularity, scaling to previously intractable problems, such as playing complex games from pixel observations, sustaining conversations with humans, and controlling robotic agents. However, there is still a wide range of domains inaccessible to RL due to the high cost and danger of interacting with the environment. Offline RL is a paradigm that learns exclusively from static datasets of previously collected interactions, making it feasible to extract policies from large and diverse training datasets. Effective offline RL algorithms have a much wider range of applications than online RL, being particularly appealing for real-world applications, such as education, healthcare, and robotics. In this work, we contribute with a unifying taxonomy to classify offline RL methods. Furthermore, we provide a comprehensive review of the latest algorithmic breakthroughs in the field using a unified notation as well as a review of existing benchmarks' properties and shortcomings. Additionally, we provide a figure that summarizes the performance of each method and class of methods on different dataset properties, equipping researchers with the tools to decide which type of algorithm is best suited for the problem at hand and identify which classes of algorithms look the most promising. Finally, we provide our perspective on open problems and propose future research directions for this rapidly growing field.

## I. Introduction

**R**EINFORCEMENT learning (RL) is a powerful learning paradigm for control. In RL, an agent must learn to maximize a specified reward signal through trial and error, i.e., actively interacting with the environment by taking actions and observing the reward. With the recent success of deep learning (DL) in complex domains (e.g., natural language processing [1] and computer vision [2]), deep RL has become increasingly popular due to its ability to leverage high-capacity function approximators, allowing agents to make decisions from unstructured inputs and with minimal feature engineering [3], [4]. A lot of the progress in DL can also be attributed to the availability of large and diverse training datasets [5]. However, current deep RL methods still typically rely on active data collection to succeed, hindering their application in the real world [6].

In the online or on-policy RL settings, an agent is free to interact with the environment and must collect a new set of experiences after every update to its policy. In off-policy RL, the agent is still free to interact with the environment. However, it can update its current policy using experiences collected from any previous policies. This increases the sample efficiency of training since the agent does not have to discard all of its previous interactions and can instead maintain a buffer where old interactions can be sampled multiple times [7].

Offline RL (also known as batch RL) is a *data-driven* RL paradigm concerned with learning exclusively from static datasets of previously collected experiences [8]. In this setting, a behavior policy interacts with the environment to collect a set of experiences, which can later be used to learn a policy without further interaction. This paradigm can be extremely valuable in settings where online interaction is impractical, either because data collection is expensive or dangerous (e.g., in robotics [9], education [10], healthcare [11], and autonomous driving [12]). Even if online interaction is viable, one might still prefer to use previously collected data for improved generalization in complex domains [8]. In Fig. 1, we illustrate the key differences between each RL paradigm. While online and off-policy RL constantly interact with the environment to update their policy, offline RL learns an offline policy from a static dataset of experiences collected by a behavior policy. After learning an offline policy, one can still opt to tune the policy online, with the added benefit that their initial policy is likely safer and cheaper to interact with the environment than an initial random policy [13].

While learning from a static dataset is one of the main benefits of offline RL, it is also what makes it so challenging for existing online RL algorithms. In theory, any off-policy method could be used to learn a policy from a dataset of previously collected experiences. However, these methods often fail when exclusively working with offline data since they were devised under the assumption that further online interactions are possible, and algorithms can typically rely on these interactions to correct erroneous behavior. Finding a balance between increased generalization and avoiding unwanted

Manuscript received 11 May 2022; revised 1 November 2022, 10 January 2023, 22 January 2023, and 13 February 2023; accepted 20 February 2023. This work was supported by the Ministry of Science, Technology, and Innovation of Brazil, Federal Law 8.248 of October 23, 1991, through the PPI-Softex. The work of Rafael Figueiredo Prudencio was supported in part by CAPES and in part by QuintoAndar. The work of Esther Luna Colombini was supported by CNPq PQ-2 under Grant 315468/2021-1. *(Corresponding author: Rafael Figueiredo Prudencio.)*

Rafael Figueiredo Prudencio and Esther Luna Colombini are with the Laboratory of Robotics and Cognitive Systems, Computing Institute, University of Campinas, Campinas 13083-852, Brazil (e-mail: rafael.prudencio@gmail.com; esther@ic.unicamp.br).

Marcos R. O. A. Maximo is with the Autonomous Computational Systems Laboratory, Computer Science Division, Aeronautics Institute of Technology, São José dos Campos 12228-900, Brazil (e-mail: mmaximo@ita.br).

Color versions of one or more figures in this article are available at https://doi.org/10.1109/TNNLS.2023.3250269.

Digital Object Identifier 10.1109/TNNLS.2023.3250269

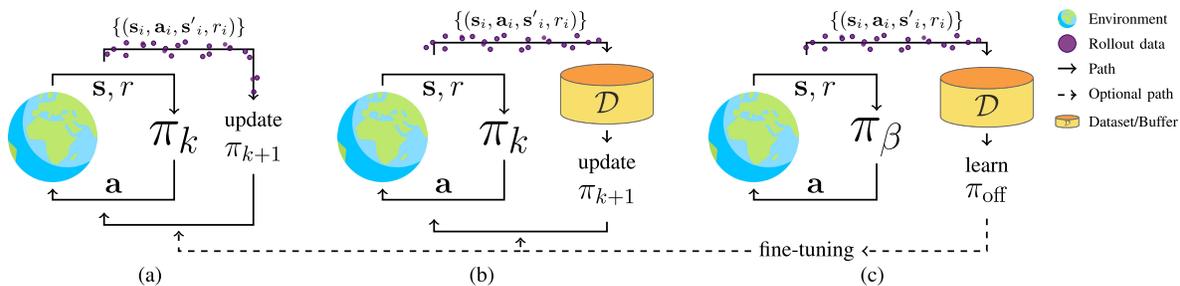

Fig. 1. Illustration of the different RL paradigms, including (a) online RL, (b) off-policy RL, and (c) offline RL. In online RL new experiences must be collected with the latest policy before making an update. In off-policy RL, we reuse previous experiences but still rely on a continuous collection of new experiences. In contrast, offline RL only uses previous experiences collected with a behavior policy $\pi_\beta$ and stored in a static dataset $\mathcal{D}$ to learn a policy $\pi_{\text{off}}$. After learning $\pi_{\text{off}}$, one can opt to fine-tune it using online or off-policy RL methods. This image is largely based on Levine et al.'s [8] pictorial illustration of RL paradigms. Earth image made by Freepik from www.flaticon.com.

behaviors outside of distribution is one of the core problems of offline RL. Moreover, this problem is further exacerbated by the widespread use of high-capacity function approximators. Most of the novel offline RL algorithms directly address this issue by proposing different losses or training procedures capable of mitigating *distributional shift.*

One of the simplest ways to address distributional shift is by directly constraining the learned policy to the behavior policy used to collect the dataset [14], [15], [16]. Other methods constrain the learned policy by making conservative estimates of future rewards, done by learning a value function that is strictly a lower bound to the true value function [17], [18]. Some model-based methods estimate the model's uncertainty using ensembles, allowing one to penalize conflicting actions and favor decisions that are consistent across the models [19], [20]. A few strategies do not explicitly restrict the learned policies but still address distributional shift by either using some variation of behavior cloning (BC) [21] or avoiding it entirely by taking a single step of policy evaluation and policy improvement [22], [23]. In contrast, some strategies do not address distributional shift at all, such as importance sampling (IS) [24], [25], [26] and trajectory optimization [27], [28], which are concerned with learning an optimal policy and an optimal trajectory distribution, respectively.

Currently, a limited number of works have reviewed the field of offline RL. Most notably, Levine et al. [8] published a tutorial article highlighting the key challenges of the field, reviewing seminal works, discussing their applications, and presenting perspectives on open problems. However, the number of offline RL and RL publications has grown exponentially over the past five years.[1] With the dramatic increase in popularity of the field, several new methods have been proposed using various strategies not discussed in Levine et al.'s [8] work. With this in mind, we were motivated to propose a novel taxonomy under a unified notation, identifying the different modules and flows that may be combined to build an offline RL algorithm. We wish to understand how these components contribute to the performance of a method and where we should focus our research in the upcoming years to advance the field further.

### A. Contributions

This survey makes the following key contributions.
1) *Taxonomy:* We propose a novel taxonomy for classifying offline RL methods. Methods can belong to different categories, including model-based, one-step, and imitation learning methods. Furthermore, methods can have several modifications to their losses, including policy constraints, regularization, or uncertainty estimation terms.
2) *Algorithmic Review:* We also provide the most updated literature review of offline RL methods using a unified notation, including detailed discussions of seminal works, recently published articles, and promising preprints.
3) *Dataset Review:* We evaluate the current benchmarks available in the literature and discuss how their datasets satisfy some of the key desirable properties of offline RL datasets. This allows fellow researchers to understand where to evaluate their method to assess its ability to address particular issues, e.g., which dataset should be used to determine if their method works well with sub-optimal data. Furthermore, it will allow one to classify their dataset according to its dataset properties and learn about common pitfalls.
4) *Method Performance*: We provide a figure with the relative performance of each method and class of methods on each of the dataset properties we defined. This will equip researchers with the tools to select the best algorithm for a given problem and identify the classes of algorithms with the most promising performance.
5) *Open Problems:* We also discuss our perspective on some of the open problems of the field, including off-policy evaluation (OPE) methods, a reliable offline RL workflow, and the ability to dynamically modulate the degree of conservatism of an algorithm.

### B. Text Organization

The rest of this survey is structured as follows. In Section II, we give a brief background of online and offline RL, introducing the notation and the key concepts behind the methods. Section III introduces our proposed taxonomy with a brief explanation of each class' structure. In Section IV, we go over each class of our taxonomy and review their main methods. Section V introduces the concept of OPE and some of the best strategies used to evaluate policies without interacting with the environment. In Section VI, we evaluate the different offline RL benchmarks available in the literature and discuss some of their properties and shortcomings. Section VIII gives an overview of the open problems in the field, suggesting promising future directions for research. Finally, Section IX presents some concluding remarks.

---

[1] A figure comparing the number of publications in offline RL and RL over the past five years is available in the Supplementary Materials. All Supplementary Materials are available at https://github.com/larocs/offline-rl-suvey

## II. Background

Toward the end of 2013, Mnih et al.'s [3] published their seminal work on deep $Q$-networks (DQN), an off-policy RL algorithm capable of learning how to play multiple Atari games at a level comparable or superior to professional game-testers only from pixel observations. Silver et al. [29] reached a new milestone in deep RL with AlphaGo, an agent trained to play Go that became the first computer program capable of beating a professional human Go player. Even with its immense progress, most deep RL applications and test environments are still limited to games and simulations [30]. Recently, to improve RL's applicability to the real world, more researchers have recognized the importance of learning from static datasets of observations. Toward the end of 2018, offline RL started to get some attention with Fujimoto et al.'s [14] work on batch-constrained $Q$-learning (BCQ), after which the field got significant traction and experienced an exponential increase in its number of publications.

In this section, we go over key concepts and ideas of offline RL. First, we define a Markov decision process (MDP) and introduce the notation used to develop the mathematical formalism of RL. Then, we formalize the problem of offline RL and detail the properties and related challenges expected from good offline RL.

### A. Markov Decision Process

An MDP is a mathematical formulation to describe an ideal environment in RL, which allows us to make theoretical statements about our problem [7]. An MDP frames the problem of learning from interactions to achieve a goal. In an MDP, an agent in a state $\mathbf{s}_t \in \mathcal{S}$ interacts with the environment by taking an action $\mathbf{a}_t \in \mathcal{A}$, and the environment responds with a new state $\mathbf{s}_{t+1} \in \mathcal{S}$ and a reward $r_t \in \mathbb{R}$, which signals how beneficial that interaction was toward the agent's goal.

The full MDP can be defined by a six-tuple $\mathcal{M} = (\mathcal{S}, \mathcal{A}, T, d_0, r, \gamma)$, where $\mathcal{S}$ denotes the state space, $\mathcal{A}$ denotes the action space, $T(\mathbf{s}_{t+1}|\mathbf{s}_t, \mathbf{a}_t)$ denotes the transition distribution, $d_0(\mathbf{s}_0)$ denotes the initial state distribution, $r(\mathbf{s}_t, \mathbf{a}_t)$ denotes the reward function, and $\gamma \in (0, 1]$ the discount factor. Within an MDP, our objective is to find a policy $\pi(\mathbf{a}_t|\mathbf{s}_t)$, which denotes the probability of taking action $\mathbf{a}_t$ conditioned on the current state $\mathbf{s}_t$. From this definition, we can derive a trajectory distribution, where a trajectory is a sequence of $H + 1$ states and $H$ actions, given by $\tau = (\mathbf{s}_0, \mathbf{a}_0, \ldots, \mathbf{s}_H)$, where $H$ may be infinite in nonepisodic environments. The probability density function for a given trajectory $\tau$ and policy $\pi$ is given by

$$p_\pi(\tau) = d_0(\mathbf{s}_0) \prod_{t=0}^{H-1} \pi(\mathbf{a}_t|\mathbf{s}_t) T(\mathbf{s}_{t+1}|\mathbf{s}_t, \mathbf{a}_t). \quad (1)$$

In an MDP, the transition distribution $T(\mathbf{s}_{t+1}|\mathbf{s}_t, \mathbf{a}_t)$ completely characterizes the environment's dynamics. In other words, the probability of a future state $\mathbf{s}_{t+1}$, depends solely on the present state $\mathbf{s}_t$ and action $\mathbf{a}_t$, without any regard for the past. The property that a state $\mathbf{s}_t$ must have all the information required to infer $\mathbf{s}_{t+1}$ after taking action $\mathbf{a}_t$ is known as the *Markov property*.

However, in most cases, instead of working with states $\mathbf{s}_t$, we have to work with observations $o_t$ of these states. When we do not have access to a fully observable state, we can define a partially observable MDP (POMDP), characterized by the eight-tuple $\mathcal{M} = (\mathcal{S}, \mathcal{A}, \mathcal{O}, T, d_0, E, r, \gamma)$, where $\mathcal{O}$ is the observation space and $E(\mathbf{o}_t|\mathbf{s}_t)$ is the emission function that maps states to observations. Within a POMDP, we wish to find a policy $\pi(\mathbf{a}_t|\mathbf{o}_t)$, conditioned on observations. Although more accurate, most offline RL works disregard the POMDP formulation and assume the Markov property is valid for the observations, which we also do throughout this survey.

### B. Reinforcement Learning

In RL, we are concerned with finding an optimal policy $\pi^*(\mathbf{a}|\mathbf{s})$ that maximizes the expected return for all trajectories induced by the policy, such that

$$\pi^* = \underset{\pi}{\mathrm{argmax}}\, \mathbb{E}_{\tau \sim p_\pi(\cdot)}[R_{0:H}] \quad (2)$$

where $R_{i:j} = \Sigma_{t'=i}^{j} \gamma^{t'-i} r(\mathbf{s}_{t'}, \mathbf{a}_{t'})$ is the discounted cumulative reward (i.e., return) of our policy from time step $i$ to $j$. For brevity, from now on, we denote $R_t = R_{t:H}$. Policy gradients are one of the key RL methods that directly maximize this objective to find $\pi^*$.

Methods like policy iteration and value iteration [7] rely on different quantities of interest to find an optimal policy, such as state-value and action-value functions. A state-value function for a policy $\pi$, denoted by $V^\pi(\mathbf{s}_t)$, maps a state to the expected return when starting from state $\mathbf{s}$ and following $\pi$ until termination, such that

$$V^\pi(\mathbf{s}_t) \doteq \mathbb{E}_{\tau \sim p_\pi(\cdot|\mathbf{s}_t)}[R_t]. \quad (3)$$

Similarly, an action-value function for a policy $\pi$, denoted by $Q^\pi(\mathbf{s}_t, \mathbf{a}_t)$, maps state-action pairs to their expected return. The difference between $Q^\pi(\mathbf{s}_t, \mathbf{a}_t)$ and $V^\pi(\mathbf{s}_t)$ is a lower variance alternative to the action-value function known as the advantage function $A^\pi(\mathbf{s}_t, \mathbf{a}_t)$ since it represents how advantageous it is to take action $\mathbf{a}_t$ as compared with the average performance we would expect from state $\mathbf{s}_t$.

These quantities are used throughout the RL field, which is conventionally subdivided into three classes of methods: dynamic programming, model free, and model based [31], [32]. Dynamic programming has its origins in optimal control [33] and may be used to compute an optimal policy based on a known MDP. In model-free methods, we assume we do not know the MDP and instead need to learn only from its samples. This can be done through policy gradients that directly learn a policy, value iteration methods that learn a value function used to extract a policy, or actor–critic methods that learn both quantities by iteratively alternating between policy evaluation and policy improvement. Finally, in model-based methods, we attempt to learn a model of the MDP, which can then be used for planning or to learn a policy by sampling from the MDP and training with a model-free approach (e.g., Dyna-based methods [34]). For a more comprehensive review of the RL field, we advise the reader to Arulkumaran et al.'s [31] or Wang et al.'s [32] deep RL survey.

### C. Offline Reinforcement Learning

A big part of modern machine learning success relies on large and diverse datasets. RL is an interactive machine learning paradigm at its core, where an agent interacts with the world, collects some experience, and uses that experience to improve its policy. Compared with other ML paradigms, we see a big gap in the generalization ability of RL, which has been successful mainly in closed and relatively narrow domains [3], [29]. The fundamental problem with an interactive learning paradigm is that every time we change the policy, we need to recollect the entire dataset, which is prohibitively expensive in the real world. One of the main reasons RL makes extensive use of simulated training [6], [35] is to avoid the cost and danger of interacting with the real-world environment.

In the offline RL setting, we have a fixed dataset collected by some unknown behavioral policy $\pi_\beta$, which is then used to learn a new and improved policy $\pi_{\text{off}}$ without further interactions with the environment. Under this paradigm, we wish to use datasets from many past experiences and generalize beyond naive imitation learning, finding, and exploiting the good parts of our behavior policy [8]. This paradigm would allow us to apply RL to domains where it is currently infeasible or impractical to collect data online, such as healthcare (e.g., medical diagnosis), robotics (e.g., robotics manipulation), inventory management, and autonomous driving. Fig. 1 depicts the main differences between the online, off-policy, and offline RL paradigms.

More formally, in the offline RL setting, we are given a static dataset of transitions $\mathcal{D} = \{(\mathbf{s}_t, \mathbf{a}_t, \mathbf{s}_{t+1}, r_t)_i\}$,[2] where $i$ indexes a transition in the dataset, the actions come from the behavior policy $\mathbf{a}_t \sim \pi_\beta(\cdot|\mathbf{s}_t)$, the states come from a distribution induced by the behavior policy $\mathbf{s}_t \sim d^{\pi_\beta}(\cdot)$, the next state is determined by the transition dynamics $\mathbf{s}_{t+1} \sim T(\cdot|\mathbf{s}_t, \mathbf{a}_t)$, and the reward is a function of state and action $r_t = r(\mathbf{s}_t, \mathbf{a}_t)$. In offline RL, the objective is still the same as in the online case: to find a policy that maximizes the expected return. However, we cannot evaluate this objective under an arbitrary trajectory distribution $p_\pi(\tau)$, since $\pi$ might experience distributional shift and visit states that we do not have any information for from our static dataset.

In Sections II-C1 and II-C2, we outline some of the desirable properties and challenges of offline RL algorithms considering their inability to further interact with the environment.

*1) Desirable Properties:* The performance of offline RL methods is often compared with a BC baseline, which tries to mimic $\pi_\beta$ from $\mathcal{D}$ in a supervised manner. Given that offline RL only has access to a static dataset, we do not have the same optimality guarantees we have in the online setting, where we are free to explore any region of our state and action spaces. However, there are still reasons why using the RL formalism to learn a new policy $\pi_{\text{off}}$ can be more beneficial than naive supervised learning. Next, we list these reasons and explain why they are desirable in a good offline RL method.

---

[2] We also refer to the transitions in $\mathcal{D}$ as $(\mathbf{s}, \mathbf{a}, \mathbf{s}', r)_i$ interchangeably. Although we represent $\mathcal{D}$ as a dataset of transitions, the dataset is often composed of a set of trajectories.

*a) Generalization:* Good behavior in one place may suggest good behavior in another place. With offline RL, we may use more extensive and diverse datasets that allow for better generalization.

*b) Filtering:* Even if the dataset is full of good and bad behaviors, finding the good ones would already result in offline RL finding a better policy than the one used to generate the data. Although selecting the good trajectories and discarding the bad ones might seem simple, this task is far from trivial in a stochastic setting. Differentiating between good and lucky behaviors is complex, and RL allows us to reason about the long-term consequences of our actions in expectation more easily through value functions, for instance.

*c) Compositionality:* Parts of good behaviors can be recombined, such that even if you have not seen good behavior in a full trajectory, you may have seen parts of good behaviors in different trajectories that can be *stitched* together.

*2) Challenges:* The most apparent reason why offline RL is difficult is its inability to interact with the environment, i.e., explore new states and experiment with new actions to find high-reward regions. Suppose $\mathcal{D}$ does not have transitions in high rewards regions. In this case, it may be impossible to learn a policy that can find such regions. Essentially, when we use a learned model or policy to act, we will inevitably see different things from what we trained on. Once the agent finds a novel state outside of the training distribution, it will make bigger mistakes that compound until the policy diverges wildly from the one it was trained on. This behavior is a type of *distributional shift*.

Here, we formalize the concept of distributional shift for offline RL. Let our objective minimize the Bellman error derived from the action-value Bellman equation. Under the offline RL setting, this gives us

$$J(\phi) = \mathbb{E}_{\mathbf{s},\mathbf{a},\mathbf{s}' \sim \mathcal{D}}\Big[\big(r(\mathbf{s},\mathbf{a}) + \gamma \mathbb{E}_{\mathbf{a}' \sim \pi_{\text{off}}(\cdot|\mathbf{s})}\big[Q_\phi^\pi(\mathbf{s}', \mathbf{a}')\big] \\ - Q_\phi^\pi(\mathbf{s}, \mathbf{a})\big)^2\Big] \quad (4)$$

where $\phi$ are the parameters of our learned $Q$-function. When minimizing $J(\phi)$, we can only expect the objective to be accurate when $\pi_\beta(\mathbf{a}|\mathbf{s}) = \pi_{\text{off}}(\mathbf{a}|\mathbf{s})$, since only then can we ensure that the $Q$-function was trained under the actions $\mathbf{a}'$ that it is being evaluated on. In practice, this should never be true since we wish to find a new policy $\pi_{\text{off}}$ that is better than our behavioral policy $\pi_\beta$, leading us to inevitably experience distributional shift under our actions. Furthermore, even if $\pi_{\text{off}}$ is able to accurately evaluate the objective on training data, the policy-induced state distribution might still deviate due to compounding errors from sampling or function approximation, i.e., $d^{\pi_{\text{off}}}(\mathbf{s}) \neq d^{\pi_\beta}(\mathbf{s})$. These errors are much more severe in offline RL since we cannot correct them through continuous interaction like in off-policy RL.

## III. TAXONOMY

In this section, we present our taxonomy for offline RL. Our objective is to devise a categorization that encompasses all offline RL methods and allows one to easily make design decisions about what to learn and how to learn it. However, this is a challenge in offline RL since several methods propose changes that are not exclusive and could be combined to form a new type of offline RL algorithm. Taxonomies are normally represented as is-a relationships (e.g., a dog is a mammal), and mereologies study has-a relationships (e.g., a dog has a tail). Therefore, we propose a high-level taxonomy that allows us to classify all offline RL algorithms. However, we still rely on mereology to distinguish different elements of each algorithm.

In Fig. 2, we illustrate our offline RL taxonomy. At a high-level, algorithms rely on an optionally filtered static dataset of transitions $\mathcal{D}$ to either learn a dynamics model, learn a trajectory distribution, or be used directly in a model-free approach to learn a policy. Both the dynamics model and trajectory distributions can be used for planning, where we use either the trajectory distribution induced by the learned dynamics model $p_{\psi_T}(\tau)$ or the trajectory distribution learned directly from the dataset $p_{\pi_\beta}(\tau)$, respectively, to determine the best set of actions to take at each given time step. However, the dynamics model can also be used to rollout extra interactions, which can then be used to learn a policy $\pi_\theta(\mathbf{a}|\mathbf{s})$. When learning a policy, we have two main algorithms: actor–critic and imitation learning methods. In actor–critic methods, we choose between one-step and multistep methods, that modify the number of policy evaluation and policy improvement steps in the algorithm. In imitation learning, we just mimic the remaining samples from our behavior policy to learn an optionally conditional policy.

The remaining modifications that can be made to offline RL algorithms we consider to be has-a relationships and illustrate them in Fig. 3. The idea is that one can optionally add any of the loss terms in the diagram to either the policy evaluation or policy improvement step of their actor–critic method. Although not illustrated in the diagram, the uncertainty estimation loss term is also used in model-based approaches to capture the uncertainty between an ensemble of dynamics models.

Finally, Table I gives an overview of the main types of modifications that can be made in an offline RL algorithm, listing their changes and extra requirements needed to implement each one. These changes are shown considering a vanilla model-free multistep actor–critic method. Sections III-A–III-H go more in-depth on the general formulations of each of these changes, leaving the literature review to Section IV.

### A. Policy Constraints

Policy constraint methods can be subdivided into direct and implicit policy constraints. Direct methods estimate the behavioral policy $\pi_\beta$ and constrain the learned policy $\pi_\theta$ to stay close to $\pi_\beta$. Implicit methods do not rely on the estimation of $\pi_\beta$ and implicitly constrain $\pi_\theta$ by using a modified objective and relying strictly on samples from $\pi_\beta$.

More formally, direct policy constraint methods address the distributional shift problem by modifying the unconstrained policy improvement objective to the constrained objective we wish to maximize

$$J(\theta) = \mathbb{E}_{\mathbf{s} \sim d^{\pi_\theta}(\cdot), \mathbf{a} \sim \pi_\theta(\cdot|\mathbf{s})} [Q^\pi(\mathbf{s}, \mathbf{a})]$$
$$\text{s.t. } D(\pi_\theta(\cdot|\mathbf{s}), \hat{\pi}_\beta(\cdot|\mathbf{s})) \le \epsilon \quad (5)$$

where $D(\cdot, \cdot)$ is some divergence metric that measures the distance between two probability distributions

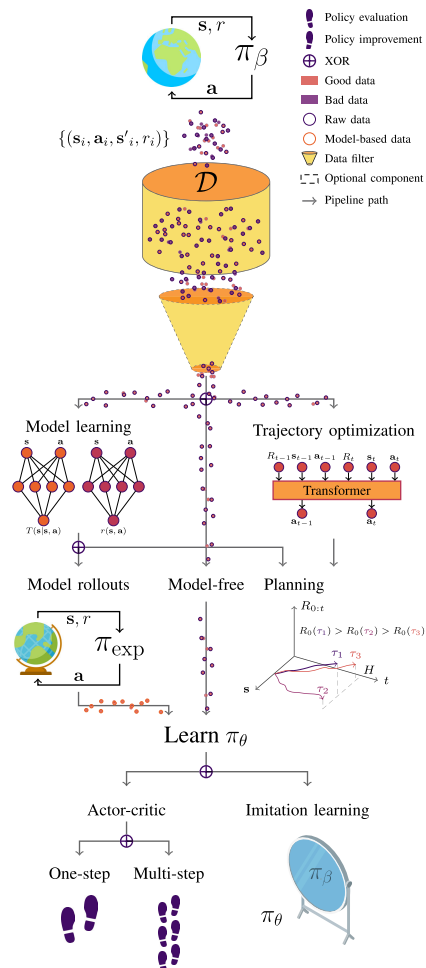

Fig. 2. Illustration of the general structure of an offline RL algorithm, where different paths represent possible algorithm design choices. Initially, the behavior policy $\pi_\beta$ interacts with the environment to collect experiences for dataset $\mathcal{D}$. The data are then optionally filtered to retain only experiences from high-return trajectories. The remaining samples can then be used to either directly learn a policy $\pi_\theta$, learn a dynamics model $T_{\psi_T}(\mathbf{s}|\mathbf{s}, \mathbf{a})$, or a model of the trajectory distribution $p_{\pi_\beta}(\tau)$. Trajectory distributions are used for planning, while dynamics models can be used either for planning or to generate synthetic samples to learn a policy. To learn $\pi_\theta$, we can opt between actor–critic and imitation learning methods. In the former, one can opt to use single or multiple steps of policy evaluation and policy improvement, while the latter typically relies on a good filtering process. Earth and globe images made by Freepik from www.flaticon.com.

(i.e., $f$-divergence), $\epsilon$ is a divergence threshold we must satisfy, and $\hat{\pi}_\beta(\cdot|\mathbf{s})$ is an estimate of the behavior policy [8]. These methods are referred to as direct methods since they directly estimate the behavior policy $\pi_\beta$ in order to compute the divergence $D$ and enforce the constraint.

This reliance on an estimate of $\pi_\beta$ is also one of the main limitations of these methods. The behavior policy can come from human-provided data, hand-designed controllers, or multiple policies, making it difficult to estimate. A few ways to estimate $\pi_\beta$ include training a parametric model with behavioral cloning (e.g., maximum likelihood over $\mathcal{D}$ [16]) or using a nonparametric naive estimator [17], such as $\hat{\pi}_\beta(\mathbf{a}|\mathbf{s}) = (\Sigma_{\mathbf{s},\mathbf{a} \in \mathcal{D}} \mathbb{1}[\mathbf{s} = \mathbf{s}, \mathbf{a} = \mathbf{a}])/(\Sigma_{\mathbf{s} \in \mathcal{D}} \mathbb{1}[\mathbf{s} = \mathbf{s}])$. However, suppose the behavior policy is incorrectly estimated, such as when we fit a unimodal policy into multimodal data. In that case, policy constraint methods can fail dramatically.

TABLE I

SUMMARY OF OFFLINE RL MODIFICATION TYPES CONSIDERING A VANILLA MODEL-FREE MULTISTEP ACTOR–CRITIC METHOD. SOME MODIFICATIONS CHANGE THE ALGORITHM'S CLASS IN TAXONOMY BY MODIFYING THEIR IS-A RELATIONSHIPS (E.G., ONE-STEP). OTHER MODIFICATIONS (E.G., POLICY CONSTRAINTS) CAN BE SIMULTANEOUSLY PRESENT IN A METHOD AND MODIFY ITS HAS-A RELATIONSHIPS

| Method Type | Modifies | Modification Type | Extra Requirements |
|---|---|---|---|
| Policy constraints | Learned policy | Probabilistic or support constraints | Estimate of $\pi_\beta$ (except implicit constraints) |
| Importance sampling | Policy gradient | Importance weights | Estimate of the product of importance weights $w_{i:j}$ or state-marginal ratio $\rho^\pi$ |
| Regularization | $Q$-function or policy | Modified $Q$-function or policy training objective | Define a regularizer |
| Uncertainty estimation | $Q$-function or model | Subtract an uncertainty | Estimate a model and define an uncertainty metric |
| Model-based | Reward function | Subtract an uncertainty | Define an uncertainty metric |
| One-step | Number of steps | Policy evaluation operator | Select a policy evaluation operator capable of dynamic programming |
| Imitation learning | Training data | Filter out bad behaviors from dataset or learn a conditional policy | Define a filter to select good behaviors or the outcomes $\omega \sim g(\cdot \mid \tau_{t:H})$ to condition the policy |
| Trajectory optimization | Trajectory distribution | Estimate entire trajectories | Select a policy search mechanism |

Another issue with policy constraints is that these methods can often be too pessimistic, which is always undesirable. For instance, if we know that a certain state has all actions with zero reward, we should not care about constraining the policy in this state once it can inadvertently affect our neural network approximator while forcing the learned policy to be close to the behavior policy in this irrelevant state. We effectively limit how good of a policy we can learn from our dataset by being too pessimistic.

Implicit policy constraint methods enforce a constraint on the learned policy $\pi_\theta$ while avoiding the need to estimate $\pi_\beta$. We can derive a solution to the constrained optimization problem in (5) by enforcing the Karush-Kuhn-Tucker (KKT) conditions [36], such that the Lagrangian is

$$\mathcal{L}(\pi, \lambda) = \mathbb{E}_{\mathbf{s} \sim d^{\pi_\beta}(\cdot)}\big[\mathbb{E}_{\mathbf{a} \sim \pi(\cdot|\mathbf{s})}\big[\hat{A}^\pi(\mathbf{s}, \mathbf{a})\big] \\ + \lambda\big(\epsilon - D_{\mathrm{KL}}\big(\pi(\cdot|\mathbf{s}) \| \pi_\beta(\cdot|\mathbf{s})\big)\big)\big] \quad (6)$$

where $\hat{A}^\pi(\mathbf{s}, \mathbf{a})$ is an estimate of the advantage function. Solving the Lagrangian for $\partial \mathcal{L}/\partial \pi = 0$ allows us to obtain a closed form nonparametric solution $\pi^*(\mathbf{a}|\mathbf{s}) \propto \pi_\beta(\mathbf{a}|\mathbf{s})\exp(\lambda^{-1}\hat{A}^{\pi_k}(\mathbf{s}, \mathbf{a}))$. Since we use parametric function approximators to estimate $\pi_\theta$, we need to project our nonparametric solution $\pi^*$ onto our policy space. One way to do this is by minimizing the Kullback-Leibler (KL) divergence between $\pi_\theta$ and $\pi^*$ in expectation under the state marginal of our data distribution. This allows us to derive the objective for the policy improvement step we wish to maximize

$$J(\theta) = \mathbb{E}_{\mathbf{s},\mathbf{a} \sim \mathcal{D}}\left[\log \pi_\theta(\mathbf{a}|\mathbf{s})\exp\left(\frac{1}{\lambda}\hat{A}^\pi(\mathbf{s}, \mathbf{a})\right)\right]. \quad (7)$$

Notice how this amounts to a weighted maximum likelihood, where the weights are given by the exponentiated advantage function. Furthermore, in order to perform the update, we do not need to learn a behavior policy explicitly and can simply use samples $(\mathbf{s}, \mathbf{a})$ from our static dataset $\mathcal{D}$.

Policy constraints are typically a loss term in the policy evaluation or improvement step of actor–critic methods. They come in two forms: distribution or support matching constraints. Distribution constraints restrict $\pi_\theta$'s distribution to match $\pi_\beta$. In contrast, support constraints only restrict the actions selected from $\pi_\theta$ to be within the support of the actions selected by $\pi_\beta$, but not the probabilities of these actions.[3] Fig. 3 illustrates the difference between each type of constraint.

### B. Importance Sampling

IS is commonly used in RL to compute off-policy policy gradients. Here, we formalize IS for offline RL as a means to evaluate our policy $\pi_\theta$ with samples from our behavior policy $\pi_\beta$. We have that the importance-weighted policy gradient $\nabla_\theta J(\theta)$ from online RL in offline RL notation is

$$\mathbb{E}_{\tau \sim p_{\pi_\beta}(\cdot)}\left[w_{0:H}\sum_{t=0}^{H}\nabla_\theta \gamma^t \log \pi_\theta(\mathbf{a}_t|\mathbf{s}_t)\hat{Q}(\mathbf{s}_t, \mathbf{a}_t)\right] \quad (8)$$

where $\hat{Q}(\mathbf{s}_t, \mathbf{a}_t)$ is our estimated expected return for $(\mathbf{s}_t, \mathbf{a}_t)$ and $w_{i:j}$ is the product of importance weights. One of the main issues with IS is that $w_{0:H}$ is exponential in $H$, making it important to devise different strategies to reduce the variance from the importance weights.

### C. Regularization

There are times when we want to impose behaviors on our learned policy that do not depend on $\pi_\beta$. The regularization is a powerful tool that allows us to tune our learned function by adding a penalty term. Let us denote by $\mathcal{R}$ our regularization term. With policy regularization, we can rewrite our policy gradient objective as

$$J(\theta) = \mathbb{E}_{\mathbf{s},\mathbf{a} \sim \mathcal{D}}\big[Q^{\pi_\theta}(\mathbf{s}, \mathbf{a})\big] + \mathcal{R}(\theta) \quad (9)$$

which we wish to maximize.

With value regularization, we penalize our learned value function to make its estimates more conservative, which gives us the modified value objective that we wish to minimize

$$J(\phi) = \mathbb{E}_{\mathbf{s},\mathbf{a},\mathbf{s}' \sim \mathcal{D}}\big[\big(r(\mathbf{s}, \mathbf{a}) + \gamma \mathbb{E}_{\mathbf{a}' \sim \pi_{\mathrm{off}}(\cdot|\mathbf{s})}\big[Q_\phi^\pi(\mathbf{s}', \mathbf{a}')\big] \\ - Q_\phi^\pi(\mathbf{s}, \mathbf{a})\big)^2\big] + \mathcal{R}(\phi). \quad (10)$$

Regularization terms tend to be less conservative than policy constraints since we do not limit our policy to $\pi_\beta$. Typically these terms are accompanied by other techniques, such as conservative models or policy constraints that effectively prevent us from taking out-of-distribution (OOD) actions. In Fig. 3, we show how regularization terms can help enforce desirable properties in our learned quantities, such as stochastic policies and conservative value functions.

---

[3]The support of a function $f : X \to \mathbb{R}$ is the set of points in its domain where $f$ is nonzero, i.e., $\mathrm{supp}(f) = \{x \in X : f(x) \neq 0\}$.

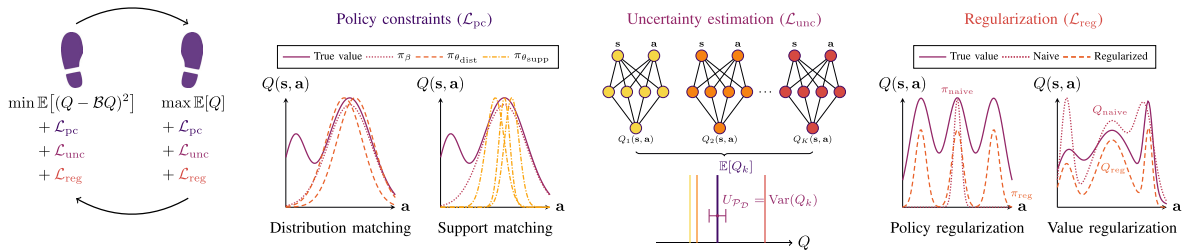

Fig. 3. Policy evaluation and policy improvement loss terms, including the policy constraints loss $\mathcal{L}_{\text{pc}}$, uncertainty estimation loss $\mathcal{L}_{\text{unc}}$, and regularization loss $\mathcal{L}_{\text{reg}}$. The policy constraints diagram showcases the difference between distribution and support matching constraints. Distribution constraints restrict $\pi_\theta$'s distribution to match $\pi_\beta$. In contrast, support constraints only restrict the actions selected from $\pi_\theta$ to be within the support of the actions selected by $\pi_\beta$. In the uncertainty estimation diagram, we show how an uncertainty measure can be extracted from the variance of predictions from an ensemble of $Q$ functions. In the regularization diagram, we illustrate how policy regularization can improve the stochasticity of the learned policy (e.g., entropy regularization), and value regularization can ensure we do not overestimate $Q$-values.

### D. Uncertainty Estimation

Uncertainty-based offline RL methods allow us to switch between conservative and naive off-policy RL methods, based on how much we trust the generalization ability of our models. By estimating the uncertainty of our approximation (e.g., policy, value function, or model), we can relax the constraints on our learned policy in low-uncertainty regions.

One of the ways to define an uncertainty estimate is with respect to our $Q$-function. Let $\mathcal{P}_\mathcal{D}(Q^\pi)$ denote the distribution of $Q$-functions for a dataset $\mathcal{D}$. We can rewrite our policy gradient objective penalizing the uncertainty captured by $\mathcal{P}_\mathcal{D}(Q^\pi)$, such that

$$J(\theta) = \mathbb{E}_{\mathbf{s},\mathbf{a}\sim\mathcal{D}}\big[\mathbb{E}_{Q^\pi\sim\mathcal{P}_\mathcal{D}(\cdot)}\big[Q^\pi(\mathbf{s},\mathbf{a})\big] - \alpha U_{\mathcal{P}_\mathcal{D}}(\mathcal{P}_\mathcal{D}(\cdot))\big] \quad (11)$$

where $U_{\mathcal{P}_\mathcal{D}}(\cdot)$ is an uncertainty measure for $\mathcal{P}_\mathcal{D}$.

Uncertainty estimation methods are typically concerned with defining the distribution $\mathcal{P}_\mathcal{D}(\cdot)$ and the uncertainty estimator $U_{\mathcal{P}_\mathcal{D}}(\cdot)$ for this distribution, which is needed to evaluate the objective. In Fig. 3, we showcase how the variance between the different $Q$-value estimates from an ensemble of $Q$-networks can be used as an uncertainty measure.

### E. Model-Based Methods

Similar to online model-based methods, offline model-based algorithms are concerned with first estimating the transition dynamics $T_{\psi_T}(\mathbf{s}_{t+1}|\mathbf{s}_t,\mathbf{a}_t)$ and the reward function $r_{\psi_r}(\mathbf{s}_t,\mathbf{a}_t)$. These functions are typically estimated using standard supervised regression with the dataset $\mathcal{D}$. We can then use the dynamics and rewards model as proxies of the real environment, simulating transitions and then using them for planning. Model-based methods often work well when the data distribution has high coverage since it is easy to learn an accurate model on this data.

Contrary to online RL, models learned offline cannot correct their mistakes by interacting with the environment. One of the ways to avoid model distributional shifts is to estimate a conservative model that avoids transitioning to OOD states. This can be done by using uncertainty estimation from Section III-D and penalizing our model's reward function in these OOD states, such that

$$\tilde{r}_{\psi_r}(\mathbf{s},\mathbf{a}) = r_{\psi_r}(\mathbf{s},\mathbf{a}) - \lambda U_r(\mathbf{s},\mathbf{a}) \quad (12)$$

where $U_r(\cdot,\cdot)$ is our state-action-dependent uncertainty measure, which we expect to be low for states and actions present in $\mathcal{D}$ and high otherwise. For examples of $U_r$, refer to Section IV-D.

### F. One-Step Methods

Most of the methods we have covered until now use actor–critic formulations to learn both a policy $\pi_\theta(\mathbf{a}|\mathbf{s})$ and an action-value function $Q_\phi^\pi(\mathbf{s},\mathbf{a})$. The actor–critic methods are normally implemented iteratively, alternating between policy evaluation and policy improvement steps in rapid succession. One of the issues with iteratively performing policy evaluation is that we inevitably run into a distributional shift, as described in Section II-C2 since we compute the target values concerning actions from our updated policy $\pi_{\text{off}}$ and train our $Q$-function on actions from our behavior policy $\pi_\beta$.

It is important to distinguish between a step and an iteration in policy improvement. An iteration consists of a single update to our quantity of interest. Meanwhile, we perform multiple iterations within a step until this quantity converges. Recent methods avoid iteratively performing policy evaluation and instead perform a single step of policy evaluation followed by a single-policy improvement step. One-step or single-step methods perform multiple state sweeps to learn an accurate estimate of $Q^{\pi_\beta}(\mathbf{s},\mathbf{a})$, as opposed to most multistep methods that continuously alternate between policy evaluation and policy improvement until $Q^{\pi_{\text{off}}}(\mathbf{s},\mathbf{a})$ converges. With an accurate estimate of $Q^{\pi_\beta}(\mathbf{s},\mathbf{a})$, one-step methods then perform a single-policy improvement step to find the best possible policy. This means we never perform policy evaluation with actions outside of our data distribution. Hence, we avoid adding constraints to our loss functions since we do not have to worry about taking OOD actions.

### G. Imitation Learning

Imitation learning[4] consists of a class of algorithms that, at their core, mimic the behavior policy $\pi_\beta$. In its simplest form, we have a BC method that exactly copies the behavior policy. This can be accomplished through supervised learning techniques, where the difference between the learned policy and behavior policy is minimized concerning some metric, i.e., we wish to minimize the objective

$$J(\theta) = D\big(\pi_\beta(\cdot|\mathbf{s}), \pi_\theta(\cdot|\mathbf{s})\big) \qquad (13)$$

where $D(\cdot, \cdot)$ is an $f$-divergence (e.g., cross-entropy). However, we have already seen in Section II-C2 how this type of objective is subject to distributional shift since tiny mistakes in our policy will inevitably lead us to query unseen states and experience compounding errors. Despite this, BC can still be successful if most of the data consists of expert behavior.

---

[4]Imitation learning methods are often classified as ones that do not make use of the reward signal to learn a policy. We extend this category to include any method that uses BC at its core, either through filtering undesirable behaviors or learning a conditional policy.

We often do not have access to an expert behavior policy in offline RL. Therefore, imitation learning methods look to filter out the bad behaviors in the dataset and only mimic the good ones. To do so, these methods often use value functions and heuristics to select only the good trajectories from the dataset to train on.

Another strategy that does not require expert behavior is to learn a conditional policy $\pi_\theta(\mathbf{a}_t|\mathbf{s}_t, \omega)$, where $\omega$ is an outcome conditioned on the remaining trajectory, i.e., $\omega \sim g(\cdot|\tau_{t:H})$ and $\tau_{i:j} = (\mathbf{s}_i, \mathbf{a}_i, \ldots, \mathbf{s}_j)$ denotes a fragment of the trajectory. By defining $g$ and learning a conditional behavior policy through BC, one can learn a powerful policy entirely offline. More formally, we wish to maximize the objective

$$J(\theta) = \mathbb{E}_{\substack{\tau \sim p_{\pi_\beta}(\cdot), t \sim \mathcal{U}(1,H) \\ \omega \sim g(\cdot|\tau_{t:H})}} \big[\log \pi_\theta(\mathbf{a}_t|\mathbf{s}_t, \omega)\big] \qquad (14)$$

where $\mathcal{U}(\cdot, \cdot)$ represents a discrete uniform distribution. One of the key challenges of these methods is defining the appropriate outcome function $g$ we wish to condition our policy on, which varies depending on the type of data and task at hand.

### H. Trajectory Optimization

In trajectory optimization, we are concerned with training a joint state-action model over entire trajectories, given by

$$p_{\pi_\beta}(\tau) = p_{\pi_\beta}(\mathbf{s}_0, \mathbf{a}_0, \ldots, \mathbf{s}_H). \qquad (15)$$

In other words, we wish to learn a model of the trajectory distribution induced by our behavior policy $\pi_\beta$. With a good model, we can then plan an optimal set of actions from an initial state $\mathbf{s}_0$.

Using a sequence modeling objective makes us less prone to selecting OOD actions. This occurs because multiple state and action anchors throughout the trajectory prevent us from deviating too far from $\pi_\beta$. Furthermore, the large models that are required to train long sequence models (e.g., transformers [37]) can work well in the offline RL setting since we can avoid active data collection and update our model between trials.

## IV. ALGORITHMIC REVIEW

In this section, we will discuss some of the recent developments in offline RL. Here, we will go over the main methods for each modification type defined in Section III. In Table II, we classify the key offline RL methods under our taxonomy. Although this classification is far from exhaustive, it gives us a good view of the most popular methods and provides insight into what areas are currently underexplored. Furthermore, in Fig. 4, we present a timeline with the key developments in the field, allowing one to evaluate the novelty and popularity of each class of methods.

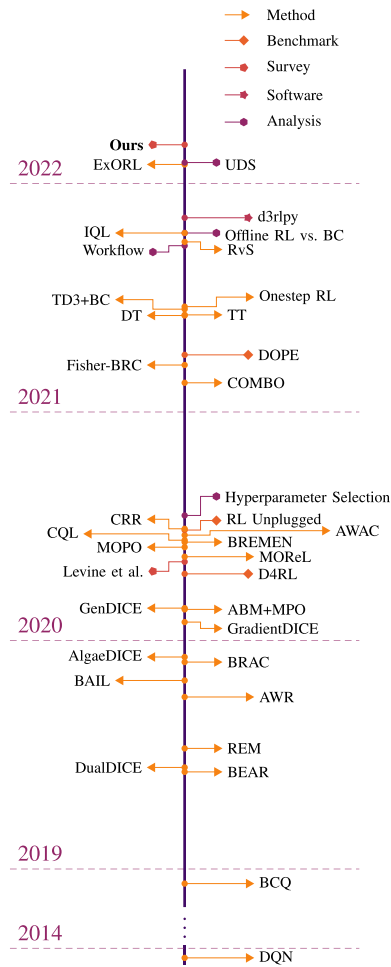

Fig. 4. Timeline illustrating the key developments, by order and interval, in the field of offline RL from the end of 2018 to the present day (2022). The timeline includes methods, benchmarks, surveys, software, and analysis papers published in the field. At the start of the timeline, we highlight DQN [3], an off-policy RL method that proposed an agent capable of learning how to play multiple Atari games from pixel observations and pioneered the field of deep RL. We then jump toward the end of 2018, highlighting BCQ [14], one of the seminal works in the field of offline RL that formally introduced some of its challenges (e.g., distributional shift). The dates shown in the timeline are the submission dates of the preprints, accessible by clicking on each event.

### A. Policy Constraints

One of the first policy constraint methods in offline RL was BCQ [14], which uses a direct policy constraint, forcing $\pi_\theta$ to be close to $\pi_\beta$ with a specific parameterization

$$\pi_\theta(\mathbf{a}, \mathbf{s}) = \operatorname*{argmax}_{\mathbf{a}_i + \xi_\theta(\mathbf{s}, \mathbf{a}_i)} Q^\pi_\phi(\mathbf{s}, \mathbf{a}_i + \xi_\theta(\mathbf{s}, \mathbf{a}_i)),$$

$$\text{for } \mathbf{a}_i \sim \hat{\pi}_\beta(\cdot|\mathbf{s}), \quad i = 1, \ldots, N \qquad (16)$$

where $\hat{\pi}_\beta(\cdot|\mathbf{s})$ is estimated using a parametric generative model trained with supervised regression, $\xi_\theta(\mathbf{s}, \mathbf{a})$ is a perturbation model with outputs bound to a predetermined range $[-\Phi, \Phi]$, and $N$ is the number of samples. BCQ constrains $\pi_\theta$ by making sure that it only chooses actions similar to what $\pi_\beta$ would choose. By using a single sample $N = 1$ and ignoring the perturbation model (i.e., $\Phi = \mathbf{0}$), we are effectively restricting $\pi_\theta = \pi_\beta$. As $N \to \infty$ and $\Phi = \mathbf{a}_{\max}$, we have an unconstrained problem similar to what is used in online RL. Although BCQ does not fit the divergence formulation we presented in (5), it can still be considered a direct policy constraint method since it relies on the estimation of $\pi_\beta$ and constrains $\pi_\theta$ to be close to $\pi_\beta$.

TABLE II
CLASSIFICATION OF DIFFERENT METHODS' MODIFICATION TYPES. IN THE TOP SECTION, WE USE A + SYMBOL TO SIGNAL THE TYPES OF HAS-A RELATIONSHIPS OF EACH METHOD. IN THE BOTTOM SECTION, WE USE A ✓ TO SIGNAL THE CLASS OF METHODS EACH METHOD BELONGS TO, WHERE NO CHECKMARKS DEFAULT TO A VANILLA MODEL-FREE MULTISTEP ACTOR–CRITIC METHOD

| Classification | | BCQ [14] | BEAR [15] | BRACp [16] | BRACv [16] | Fisher-BRC [38] | CQL [17] | AWR [36] | AWAC [39] | COMBO [18] | WIS [40] | DRIS [41] | AlgaeDICE [25] | GenDICE [24] | DualDICE [26] | Onestep RL [22] | IQL [23] | MOPO [42] | MOReL [20] | BREMEN [43] | TD3+BC [44] | BAIL [21] | CRR [45] | ABM-MPO [46] | RvS [47] | DT [27] | TT [28] |
|---|---|---|---|---|---|---|---|---|---|---|---|---|---|---|---|---|---|---|---|---|---|---|---|---|---|---|---|
| Policy Constraints | Direct | + | | + | + | | | | | | | | | | | | | | | | + | | | | | | |
| | Implicit | | + | | + | | + | + | | | | | | | | | + | + | | | | + | | + | | | |
| Regularization | Policy | | | + | | | + | | + | | | | | | | | | | | | | | | | | | |
| | Value | | | | + | + | | | + | | | | | + | | | | + | | | | | | | | | |
| Uncertainty Estimation | | | | | | | | | | | | | | | | | | | + | + | | | | | | | |
| Importance Sampling | Weighted | | | | | | | | | | + | | | | | | | | | | | | | | | | |
| | Doubly-robust | | | | | | | | | | | + | | | | | | | | | | | | | | | |
| | Marginalized | | | | | | | | | | | | + | + | + | | | | | | | | | | | | |
| One-step | | | | | | | | | | | | | | | | ✓ | ✓ | | | | | | | | | | |
| Model-based | | | | | | | | | | ✓ | | | | | | | | ✓ | ✓ | ✓ | | | | | | | |
| Imitation Learning | | | | | | | | | | | | | | | | | | | | | | ✓ | ✓ | ✓ | ✓ | ✓ | |
| Trajectory Optimization | | | | | | | | | | | | | | | | | | | | | | | | | | ✓ | ✓ |

After BCQ, Kumar et al. [15] argued that constraining $\pi_\theta$ to match $\pi_\beta$'s distribution would limit how good of a policy we could learn since we would not be able to exploit the good actions in $\pi_\beta$ and ignore the poor ones. Kumar et al. [15] distinguish between distribution and support matching divergences, arguing that support matching is superior since we still ensure that we do not take OOD actions in the learned policy while not restricting ourselves to copying the poor behaviors in $\pi_\beta$. They propose a novel method named bootstrapping error accumulation reduction (BEAR), which uses the maximum mean discrepancy (MMD) divergence $D_{\text{MMD}}$ with a Gaussian kernel as the $f$-divergence in (5) to constrain $\pi_\theta$. Empirically, they find that when computing $D_{\text{MMD}}$ over a small number of samples, the sampled MMD between $\pi_\beta$ and $\pi_\theta$ is similar to the MMD between $\pi_\beta$'s support and $\pi_\theta$. In their experiments, they show how BEAR can find good policies even when $\pi_\beta$ is composed of several suboptimal behaviors, while BCQ has a hard time filtering out poor behaviors. Since the sample-based MMD can be computed directly through samples from $\pi_\beta$, we classify BEAR as an indirect policy constraint method.

In succession, Wu et al. [16] proposed behavior regularized actor–critic (BRAC), a general framework for behavior-regularized actor–critic methods. BRAC allows one to penalize either the policy improvement or policy evaluation step by subtracting a divergence term from either objective. In their work, Wu et al. [16] show that many of the proposed changes from BCQ and BEAR were not significant to each method's performance. Decisions like the type of divergence to use (e.g., MMD versus KL) were far less significant than performing an extensive hyperparameter search. Overall, they found that applying a value penalty was more beneficial than regularizing the policy, where they achieved the best results with BRAC-v and BRAC-p, which use $D_{\text{KL}}(\pi_\theta \| \pi_\beta)$ as the $f$-divergence to constrain the policy in the policy evaluation step and policy improvement step, respectively. Since BRAC relies on a maximum log-likelihood estimate of $\pi_\beta$, it is considered a direct policy constraint method.

Afterward, Kostrikov et al. [38] proposed Fisher-behavior regularized critic (Fisher-BRC), which uses a Fisher divergence [48] $D_F(\pi_\theta \| \pi_\beta)$ to constrain the entropy-regularized learned policy. In the critic optimization objective, they propose using entropy-smoothed $Q$-values, such that $Q_\phi^\pi(\mathbf{s}, \mathbf{a}) = O_\phi^\pi(\mathbf{s}, \mathbf{a}) + \log \hat{\pi}_\beta(\mathbf{s}, \mathbf{a})$, where $O_\phi^\pi(\mathbf{s}, \mathbf{a})$ is a state-action offset function. Replacing these $Q$-values both in the TD minimization objective from (4) and the policy learning objective from (2) with entropy regularization yields the basic actor–critic formulation for Fisher-BRC. To prevent the gradient from $O_\phi^\pi$ from dominating the gradient from $\hat{\pi}_\beta$, they also add a gradient penalty $\|\nabla_\mathbf{a} O_\phi^\pi(\mathbf{s}, \mathbf{a})\|^2$ to the critic optimization objective. Although this method could be interpreted as a value regularization method, where a gradient penalty is used to regularize the state-action offset function, Kostrikov et al. [38] show that the same objective can be derived by adding a Fisher divergence term between the entropy-regularized policy (i.e., Boltzmann policy) and the behavior policy in the critic objective.

Moving away from estimating the behavior policy, Peng et al. [36] proposed the advantage-weighted regression (AWR) method (analogous to MARWIL [49]). This actor–critic algorithm implicitly applies a KL divergence constraint in the policy improvement step. AWR uses Monte Carlo rollouts to train a value function $V_\phi^\pi(\mathbf{s}_t)$ with supervised regression. In the policy improvement step, AWR uses the Monte Carlo advantage function, i.e., $\hat{A}^\pi = A_{\text{MC}}^\pi$, as the exponential weights of the weighted maximum log-likelihood objective from (7), where $A_{\text{MC}}^\pi$ is given by

$$A_{\text{MC}}^\pi(\mathbf{s}_t, \mathbf{a}_t) = R_t - V_\phi^\pi(\mathbf{s}_t). \tag{17}$$

Similar to AWR, Nair et al. [39] proposed the advantage-weighted actor–critic (AWAC) method, which uses a $Q$-function to estimate the advantage in order to reduce its variance and increase sample efficiency. The $Q$-function is fit using a bootstrapped regression, where we minimize the objective $J(\phi)$ given by

$$\mathbb{E}_{\mathbf{s},\mathbf{a},\mathbf{s}'\sim\mathcal{D}}\left[\frac{1}{2}(r(\mathbf{s},\mathbf{a}) + \gamma \mathbb{E}_{\mathbf{a}'\sim\pi_{\theta_{k-1}}(\cdot|\mathbf{s}')}[Q_\phi^\pi(\mathbf{s}',\mathbf{a}')] - Q_\phi^\pi(\mathbf{s},\mathbf{a}))^2\right]. \quad (18)$$

Notice how this is slightly different from the value-based objective, which takes the maximum $Q$-value over the next actions instead of the expectation under the latest policy. In the policy improvement step, AWAC uses the same advantage-weighted maximum log-likelihood objective from (7) as AWR, with an advantage estimate $\hat{A}^\pi$-based solely on the action-value function, that is,

$$A_{\text{AWAC}}^\pi(\mathbf{s},\mathbf{a}) = Q_\phi^\pi(\mathbf{s},\mathbf{a}) - \mathbb{E}_{\tilde{\mathbf{a}}\sim\pi_\theta(\cdot|\mathbf{s})}[Q_\phi^\pi(\mathbf{s},\tilde{\mathbf{a}})]. \quad (19)$$

However, using an action-dependent baseline requires adding an error term to the loss to correct for the bias, similar to $Q$-Prop [50]. Interestingly, Jiang and Li [39] do not attempt to correct this bias, likely because their advantage estimate would still be biased due to its reliance on a bootstrapped return.

Finally, in an effort to simplify increasingly complex offline RL methods, Fujimoto and Gu [44] propose adding a behavior-cloning regularizer to the policy improvement step of the twin delayed DDPG (TD3) algorithm [51], such that

$$J(\theta) = \mathbb{E}_{\mathbf{s},\mathbf{a}\sim\mathcal{D},\tilde{\mathbf{a}}\sim\pi_\theta(\cdot|\mathbf{s})}\left[\lambda Q_\phi^\pi(\mathbf{s},\tilde{\mathbf{a}}) - (\tilde{\mathbf{a}}-\mathbf{a})^2\right] \quad (20)$$

where $\lambda$ controls the strength of the regularizer. By penalizing the mean squared error (MSE) between actions sampled from the learned policy $\pi_\theta$ and actions sampled from $\pi_\beta$, TD3 + BC applies a form of implicit policy constraint. Using this simple method where they simply modify the TD3 [51] algorithm by applying $z$-score normalization to the states and using the BC regularizer earlier, Fujimoto and Gu [44] achieve competitive results with SOTA methods on datasets for deep data-driven reinforcement learning's (D4RL's) Gym-MuJoCo suite, showing how performance is often not tied to algorithmic complexity.

Implicit policy constraint methods, such as BEAR, AWR, AWAC, and TD3+BC, are particularly promising for online fine-tuning after offline training. Direct policy constraint methods have a hard time reestimating $\pi_\beta$ every time the dataset $\mathcal{D}$ changes as we collect more data online. By avoiding this estimation, we can seamlessly switch between an offline and online environment by simply appending the new online transitions to our offline dataset $\mathcal{D}$. In practice, according to Nair et al. [39], we see that AWR and AWAC significantly outperform BEAR and BRAC when fine-tuning online after training offline with suboptimal data.

### B. Importance Sampling

Precup et al. [40] propose some of the first strategies to mitigate the high variance in vanilla importance-weighted policy gradients given in (8). If we use a Monte Carlo return estimate, such that $\hat{Q}(\mathbf{s}_t,\mathbf{a}_t) = \sum_{t'=t}^{H}\gamma^{t'-t}r(\mathbf{s}_{t'},\mathbf{a}_{t'})$, and observe that present rewards do not depend on future states and actions, we can rewrite our per-trajectory IS policy gradient as a per-decision one, that is,

$$\nabla_\theta J(\theta) = \mathbb{E}_{\tau\sim p_{\pi_\beta}(\cdot)}\left[\sum_{t=0}^{H} w_{0:t-1}\nabla_\theta \gamma^t \log\pi_\theta(\mathbf{a}_t|\mathbf{s}_t) \cdot \sum_{t'=t}^{H} w_{t:t'}\gamma^{t'-t}r(\mathbf{s}_{t'},\mathbf{a}_{t'})\right]. \quad (21)$$

This gives us an unbiased estimator of $\nabla_\theta J(\theta)$ with lower variance since we have $w_{t:t'}$ in place of $w_{t:H}$ weighing each reward, which has a less than or equal to the number of terms being multiplied per reward. Precup et al. [40] also suggest using self-normalizing importance weights $\tilde{w}_{i:j} = w_{i:j}/\mathbb{E}_{\tau\sim p_{\pi_\beta}(\tau)}[w_{i:j}]$ in place of $w_{i:j}$ in (21), trading off additional bias for a large reduction in variance.

To further reduce variance, Jiang and Li [41] propose the doubly robust estimator, which incorporates $Q$-function estimates as control variates into the importance-sampled estimator, modifying the objective to

$$J(\pi_\theta) = \mathbb{E}_{\tau\sim p_{\pi_\beta}(\cdot)}\left[\hat{V}^{\pi_\theta}(\mathbf{s}_0) + \sum_{t=0}^{H-1} w_{0:t} \cdot (r(\mathbf{s}_t,\mathbf{a}_t) + \gamma^{t+1}\hat{V}^{\pi_\theta}(\mathbf{s}_{t+1}) - \hat{Q}^{\pi_\theta}(\mathbf{s}_t,\mathbf{a}_t))\right] \quad (22)$$

where $\hat{V}^{\pi_\theta}(\mathbf{s}_t) = \mathbb{E}_{\mathbf{a}\sim\pi_\theta(\cdot|\mathbf{s}_t)}[\hat{Q}^{\pi_\theta}(\mathbf{s}_t,\mathbf{a})]$. If we are given an estimate of $\hat{Q}^{\pi_\theta}$, possibly via regression with a different dataset, we can use this unbiased estimator to reduce the variance of our gradients. The estimator is considered doubly robust since it is unbiased if either $\pi_\beta$ is known or if $\hat{Q}^{\pi_\theta}$ is correctly estimated. Despite these variance reduction efforts, all methods shown until now still rely on the product of importance weights $w_{i:j}$ that are exponential in $H$ and make IS poorly conditioned.

To avoid these exponential weights, marginalized IS [52] uses an estimate of the state-marginal importance ratio $\rho^\pi(\mathbf{s}) = (d^\pi(\mathbf{s}))/(d^{\pi_\beta}(\mathbf{s}))$ to weigh the rewards at each time step. By definition, a state-marginal is $d^\pi(\mathbf{s}_t) = d_0(\mathbf{s}_0)\prod_{t'=0}^{t-1}\pi(\mathbf{a}_{t'}|\mathbf{s}_{t'})T(\mathbf{s}_{t'+1}|\mathbf{s}_{t'},\mathbf{a}_{t'})$. Considering that $d_0(\mathbf{s}_0)$ and $T(\mathbf{s}_{t+1}|\mathbf{s}_t,\mathbf{a}_t)$ are part of the MDP and independent of the policy, these terms cancel out in the state-marginal importance ratio, such that $\rho^{\pi_\theta}(\mathbf{s}_t) = w_{0:t}$. By estimating $\rho^{\pi_\theta}(\mathbf{s})$ directly and substituting it in (21), we eliminate the need to multiply $O(H)$ terms together, reducing the variance of our policy gradient.

Furthermore, we have that the state-marginal importance ratio satisfies the following Bellman equation:

$$d^{\pi_\beta}(\mathbf{s}')\rho^\pi(\mathbf{s}') = (1-\gamma)d_0(\mathbf{s}') + \gamma\sum_{\mathbf{s},\mathbf{a}} d^{\pi_\beta}(\mathbf{s})\rho^\pi(\mathbf{s})\pi(\mathbf{a}|\mathbf{s})T(\mathbf{s}'|\mathbf{s},\mathbf{a}) \quad (23)$$

which we can leverage to perform temporal difference updates and estimate $\rho^\pi(\mathbf{s})$ under our policy. To solve for $\rho^\pi(\mathbf{s})$, we typically minimize the difference between both sides of the equation, making sure the term $d^{\pi_\beta}(\mathbf{s})$ multiplies everything. That way, we can approximate the value using samples from our dataset $\mathcal{D}$, without the need to estimate $d^{\pi_\beta}(\mathbf{s})$.

Zhang et al. [24] propose the generalized stationary DIstribution correction estimation (GenDICE) method, which extends this constraint to state-action marginal importance ratios, $\rho^\pi(\mathbf{s}, \mathbf{a}) = (d^\pi(\mathbf{s}, \mathbf{a}))/(d^{\pi_\beta}(\mathbf{s}, \mathbf{a}))$, and directly optimize the residual error corresponding to its modified Bellman equation. GenDICE uses the constraint

$$d^{\pi_\beta}(\mathbf{s}', \mathbf{a}')\rho^\pi(\mathbf{s}', \mathbf{a}') = (1-\gamma)d_0(\mathbf{s}')\pi(\mathbf{a}'|\mathbf{s}') + \gamma \sum_{\mathbf{s},\mathbf{a}} d^{\pi_\beta}(\mathbf{s}, \mathbf{a})\rho^\pi(\mathbf{s}, \mathbf{a})\pi(\mathbf{a}|\mathbf{s})T(\mathbf{s}'|\mathbf{s}, \mathbf{a}) \quad (24)$$

and minimizes a divergence metric $D_f$ between the two sides of the equation, subject to the constraint that $\rho^\pi(\mathbf{s}, \mathbf{a})$ must integrate to unity in expectation over the dataset $\mathcal{D}$. There is a wide range of marginalized IS methods proposed for offline RL [25], [26], [53], which we will not cover in this survey for brevity and since Levine et al. [8] already do a good job of discussing.

### C. Regularization

Regarding policy regularization, Haarnoja et al. [54] proposed an entropy regularization term in their seminal work on soft actor–critics (SACs). By adding the following regularization term to the policy gradient objective from (9),

$$\mathcal{R}(\theta) = \mathbb{E}_{\mathbf{s} \sim \mathcal{D}}[\mathcal{H}(\pi_\theta(\cdot|\mathbf{s}_t))] \\ = -\mathbb{E}_{\mathbf{s} \sim \mathcal{D}, \mathbf{a} \sim \pi_\theta(\cdot|\mathbf{s})}[\log \pi_\theta(\mathbf{a}|\mathbf{s})] \quad (25)$$

one is able to control the stochasticity of the optimal policy. Adding an entropy maximization term helps improve the robustness and stability of our training procedure since it avoids premature convergence of the policy variance. The more weight we put into this regularization term, the more stochastic we wish the policy to be.

Regarding value regularization, Nachum et al. [25] introduce a term in the Bellman error objective from (10) that pushes $Q$-values down for actions sampled from the learned policy $\pi_\theta$ to avoid the overestimation of values in OOD actions, such that

$$\mathcal{R}(\phi) = \mathbb{E}_{\mathbf{s} \sim \mathcal{D}, \mathbf{a} \sim \pi_\theta(\cdot|\mathbf{s})}[Q_\phi^\pi(\mathbf{s}, \mathbf{a})]. \quad (26)$$

Similarly, Kumar et al. [17] propose an offline RL method named constrained $Q$-learning (CQL) that learns a lower bound of the true $Q$-function by adding value regularization terms to its objective. In its most general form, the CQL regularizer is given by

$$\mathcal{R}(\phi) = \max_\mu \mathbb{E}_{\mathbf{s} \sim \mathcal{D}, \mathbf{a} \sim \mu(\cdot|\mathbf{s})}[Q_\phi^\pi(\mathbf{s}, \mathbf{a})] \\ - \mathbb{E}_{\mathbf{s} \sim \mathcal{D}, \mathbf{a} \sim \hat{\pi}_\beta(\cdot|\mathbf{s})}[Q_\phi^\pi(\mathbf{s}, \mathbf{a})] + \mathcal{R}(\mu) \quad (27)$$

where $\mu(\cdot|\mathbf{s})$ is a policy that visits the unseen actions in $\mathcal{D}$ (i.e., OOD actions), $\hat{\pi}_\beta(\cdot|\mathbf{s})$ is an estimate of the behavior policy $\pi_\beta$, and $\mathcal{R}(\mu)$ is a regularization term for the policy $\mu(\mathbf{a}|\mathbf{s})$. In their work, Kumar et al. [17] show that with this regularizer, CQL learns a state-value function that strictly underestimates the values for all states in the dataset, i.e., $\forall \mathbf{s} \in \mathcal{D}, V_{\text{CQL}}(\mathbf{s}) \leq V(\mathbf{s})$.

The intuition behind this regularization term is that it will push up values that are seen in $\mathcal{D}$, possibly overestimating them, and pull down values in unseen actions. The Bellman error in value regularization objective from (10) ensures that in-distribution state-action values are accurate, while the negative expectation term in the CQL regularizer in (27) pushes these values up. The positive expectation term pulls the values for OOD actions down, while the regularization term helps shape the policy $\mu(\mathbf{a}|\mathbf{s})$ to ensure it visits these actions. The maximization term over $\mu$ ensures that $\mu(\mathbf{a}|\mathbf{s})$ approximates the policy that would maximize the current $Q$-function iterate, giving rise to an online algorithm.

One of the drawbacks of this approach is that it has a saddle point problem since we are both minimizing and maximizing the $Q$-function, which can be unstable to solve in practice. Although Kumar et al. [17] present several choices for $\mathcal{R}(\mu)$, one of the simplest options is to use $\mathcal{H}(\mu)$, such that the optimal solution to the maximization term is $\mu^* = 1/z \exp(Q^\pi(\mathbf{s}, \mathbf{a}))$, where $Z$ is a normalizing factor. Plugging this into the CQL regularizer from (27), we have that

$$\mathcal{R}(\phi) = \mathbb{E}_{\mathbf{s} \sim \mathcal{D}}\left[\log \sum_a \exp(Q_\phi^\pi(\mathbf{s}, \mathbf{a})) - \mathbb{E}_{\mathbf{a} \sim \hat{\pi}_\beta(\cdot|\mathbf{s})}[Q_\phi^\pi(\mathbf{s}, \mathbf{a})]\right] \quad (28)$$

which avoids the maximization term and offers more stability in training. Although CQL achieves good performance in the D4RL [6] benchmark, the log–sum–exp term in (28) is intractable for continuous actions and must be computed through numerical integration. Kumar et al. [17] opt to use Monte-Carlo sampling [55] with importance weights, where samples are drawn from the current training policy.

It is worth noting that this method typically outperforms policy constraint methods in several challenging tasks, including the AntMaze and Kitchen domains from D4RL [6]. These domains require algorithms to learn how to stitch suboptimal behavior, which policy constraint methods have difficulty doing. Singh et al. [56] apply CQL to a complex environment with prior image data of how to solve simple tasks like closing a drawer and picking up an object using sparse rewards. During the evaluation, Singh et al. [56] present the agent with initial states unseen during training, where it can compose different tasks learned on the prior data to close the drawer and pick up an object, for instance.

### D. Uncertainty Estimation

One way to estimate uncertainty is with an ensemble of $Q$-functions. As a naive attempt, Agarwal et al. [57] train a set of $K$ $Q$-functions $Q_{\phi_1}^\pi, Q_{\phi_2}^\pi, \ldots, Q_{\phi_K}^\pi$ independently by using disjoint partitions of the dataset for each $Q$-function. This approach allows approximating the $Q$-function distribution as

$$\mathcal{P}_\mathcal{D}(Q^\pi) \approx \frac{1}{K} \sum_{i=1}^K \delta[Q^\pi = Q_{\phi_i}^\pi] \quad (29)$$

allowing us to use sample means and sample variances for the expectation and uncertainty terms in uncertainty-based objective from (11) (i.e., $U_{\mathcal{P}_\mathcal{D}} = \text{Var}$). In practice, this ensemble offers very little diversity, causing the uncertainty to be underestimated and making us more prone to take OOD actions [58].

Agarwal et al. [57] also propose the random ensemble mixture (REM) method, where they sample a random convex combination of $Q$-functions and use them to estimate the $Q$-value for a given state-action tuple, that is,

$$Q^\pi(\mathbf{s}, \mathbf{a}) = \sum_{i=1}^{K} \alpha_i Q^\pi_{\phi_i}(\mathbf{s}, \mathbf{a})$$
$$\text{s.t.} \sum_i \alpha_i = 1 \text{ and } \forall i, \alpha_i > 0. \quad (30)$$

This method can work well in datasets with high coverage compared with a standard naive off-policy RL method, such as a DQN [3], but still lags behind other approaches like policy constraints.

### E. Model-Based Methods

Like uncertainty estimation methods, several model-based methods are also concerned with estimating uncertainty. However, model-based methods typically use an uncertainty measure to constrain their model. Kidambi et al. [20] propose a method named model-based offline RL (MOReL), which measures their model's epistemic uncertainty through an ensemble of dynamics models. They use a measure of disagreement between the dynamics given by

$$\text{dis}_{i,j}(\mathbf{s}_t, \mathbf{a}_t) = \mathbb{E}_{\substack{\mathbf{s}^i_{t+1} \sim T_{\psi_i}(\cdot|\mathbf{s}_t, \mathbf{a}_t) \\ \mathbf{s}^j_{t+1} \sim T_{\psi_j}(\cdot|\mathbf{s}_t, \mathbf{a}_t)}} \left[ \|\mathbf{s}^i_{t+1} - \mathbf{s}^j_{t+1}\| \right] \quad (31)$$

to define the uncertainty measure as

$$U_r(\mathbf{s}, \mathbf{a}) = \begin{cases} r_{\max}, & \text{if } \max_{i,j} \text{dis}_{i,j}(s, a) > \text{threshold} \\ 0, & \text{otherwise}. \end{cases} \quad (32)$$

Yu et al. [42] propose another method named model-based offline policy optimization (MOPO), which uses the maximum prediction uncertainty from an ensemble of models, where they have the model dynamics given by

$$T_{\psi_i}(\mathbf{s}'|\mathbf{s}, \mathbf{a}) = \mathcal{N}(\mu_i(\mathbf{s}, \mathbf{a}), \Sigma_i(\mathbf{s}, \mathbf{a})) \quad (33)$$

where $\mu_i$ and $\Sigma_i$ are the mean and covariance matrices of the multivariate Gaussian used to model the $i$th transition dynamics from the ensemble, respectively. The uncertainty is then defined as

$$U_r(\mathbf{s}, \mathbf{a}) = \max_i \|\Sigma_i(\mathbf{s}, \mathbf{a})\|_F \quad (34)$$

where $\|\cdot\|_F$ is the Frobenius norm.

With these modified reward functions in hand, we can then use classic model-based approaches to tackle offline RL problems, such as Dyna-based methods that sample transitions from the model to train a model-free algorithm [19], [34]. We can also run trajectory optimization or planning under the learned model with methods like linear-quadratic regulator (LQR) [59] and Monte Carlo tree search (MCTS) [60].

Concurrently, Matsushima et al. [43] proposed the behavior-REgularized model-ENsemble (BREMEN) method, which learns an ensemble of models of the behavior MDP, as opposed to a pessimistic MDP. In addition, it implicitly constrains the policy to be close to the behavior policy through trust-region policy updates [61]. BREMEN is a method akin to direct policy constraint methods since it learns an estimate of the behavior policy $\hat{\pi}_\beta$, using it to initialize the training policy $\pi_{\theta_0}$ as Gaussian policy with mean from $\hat{\pi}_\beta$ and unit variance. To update its training policy, at every iteration $j$, it samples a model from the ensemble of $K$ learned dynamics models $T_{\psi_i}(\cdot|\hat{\mathbf{s}}_t, \hat{\mathbf{a}}_t)$ and uses it to obtain trajectory rollouts with the current training policy, creating a new dataset $\hat{\mathcal{D}}_j$ with transitions $(\hat{\mathbf{s}}_t, \hat{\mathbf{a}}_t, \hat{\mathbf{s}}_{t+1}, \hat{r}_t)$, where $\hat{\mathbf{s}}_0 \sim d_0(\cdot)$, $\hat{\mathbf{a}}_t \sim \pi_{\theta_j}(\cdot|\hat{\mathbf{s}}_t)$, $\hat{\mathbf{s}}_{t+1} \sim T_{\psi_i}(\cdot|\hat{\mathbf{s}}_t, \hat{\mathbf{a}}_t)$, $\hat{r}_t = r(\hat{\mathbf{s}}_t, \hat{\mathbf{a}}_t)$, and $i \sim \mathcal{U}(1, K)$. After creating $\hat{\mathcal{D}}_j$, the policy is optimized by maximizing the objective

$$J(\theta_{j+1}) = \mathbb{E}_{\hat{\mathbf{s}}, \hat{\mathbf{a}} \sim \hat{\mathcal{D}}_j} \left[ \frac{\pi_{\theta_{j+1}}(\hat{\mathbf{a}}|\hat{\mathbf{s}})}{\pi_{\theta_j}(\hat{\mathbf{a}}|\hat{\mathbf{s}})} \hat{A}^{\pi_{\theta_j}}(\hat{\mathbf{s}}, \hat{\mathbf{a}}) \right]$$
$$\text{s.t.} \mathbb{E}_{\hat{\mathbf{s}} \sim \hat{\mathcal{D}}_j} \left[ D_{\text{KL}}(\pi_{\theta_{j+1}}(\cdot|\hat{\mathbf{s}}) \| \pi_{\theta_j}(\cdot|\hat{\mathbf{s}})) \right] \leq \epsilon \quad (35)$$

where $\hat{A}^{\pi_{\theta_j}}(\hat{\mathbf{s}}, \hat{\mathbf{a}})$ is the advantage of $\pi_{\theta_j}$ computed with model-based rollouts using the sampled dynamics model $T_{\psi_i}$ for the given iteration.

While BREMEN performs well in the offline setting against policy constraint methods that preceded it, Matsushima et al. [43] argue that its main benefit is its *deployment efficiency* when starting from a small dataset collected with a random behavior policy. Deployment efficiency measures the number of distinct data-collection policies used to train a given policy. While offline RL lies at one extreme of deployment efficiency with a single data-collection policy (i.e., $\pi_\beta$), off-policy and online RL lie at the opposite extreme with thousands or millions of different interactions with the environment. BREMEN shows good results in limited deployment settings, obtaining successful policies from initial random policies in 5–10 deployments, while the recursive application of other offline RL methods shows limited improvement in successive deployments. One caveat of this measure is that we did not find any other works that benchmark their deployment efficiency, making it difficult to compare BREMEN to other offline RL methods.

More recently, Yu et al. [18] proposed a method dubbed conservative offline model-based policy optimization (COMBO), which is a model-based version of CQL [17]. COMBO learns a single-dynamics model $T_{\psi_T}(\mathbf{s}'|\mathbf{s}, \mathbf{a})$ as a Gaussian distribution over the next state and reward trained via maximum log-likelihood. This learned dynamics model induces a new MDP, which we will denote by $\hat{\mathcal{M}}$.

In the policy evaluation step, it minimizes the following objective $J(\phi)$ given by:

$$\alpha \left( \mathbb{E}_{\mathbf{s}, \mathbf{a} \sim \rho(\mathbf{s}, \mathbf{a})} [Q^\pi_\phi(\mathbf{s}, \mathbf{a})] - \mathbb{E}_{\mathbf{s}, \mathbf{a} \sim \mathcal{D}} [Q^\pi_\phi(\mathbf{s}, \mathbf{a})] \right)$$
$$+ \frac{1}{2} \mathbb{E}_{\mathbf{s}, \mathbf{a}, \mathbf{s}' \sim d^\mu_f} \left[ \left( Q^\pi_\phi(\mathbf{s}, \mathbf{a}) - (\widehat{\mathcal{B}}^\pi Q^\pi_\phi)(\mathbf{s}, \mathbf{a}) \right)^2 \right] \quad (36)$$

where $(\widehat{\mathcal{B}}^\pi Q^\pi_\phi)(\mathbf{s}, \mathbf{a})$ is the sample-based Bellman operator,[5] and we highlight $\rho(\mathbf{s}, \mathbf{a})$ and $d^\mu_f$ in blue as they are the only things that change from the CQL evaluation step. For $\rho(\mathbf{s}, \mathbf{a})$, they choose

$$\rho(\mathbf{s}, \mathbf{a}) = d^\pi_{\hat{\mathcal{M}}}(\mathbf{s}) \pi(\mathbf{a}|\mathbf{s}) \quad (37)$$

---

[5]The Bellman operator can be defined as $(\mathcal{B}^\pi Q^\pi)(\mathbf{s}, \mathbf{a}) \doteq r(\mathbf{s}, \mathbf{a}) + \gamma \mathbb{E}_{\mathbf{s}' \sim T(\cdot|\mathbf{s}, \mathbf{a}), \mathbf{a}' \sim \pi(\cdot|\mathbf{s}')}[Q^\pi(\mathbf{s}', \mathbf{a}')]$. The sample-based Bellman operator drops the expectations and is defined as $(\widehat{\mathcal{B}}^\pi Q^\pi)(\mathbf{s}, \mathbf{a}) \doteq r(\mathbf{s}, \mathbf{a}) + Q^\pi(\mathbf{s}', \mathbf{a}')$, where $\mathbf{s}' \sim \mathcal{D}$ and $\mathbf{a}' \sim \pi(\cdot|\mathbf{s}')$.

where $d_{\hat{\mathcal{M}}}^\pi(\mathbf{s})$ is the discounted marginal state distribution when executing $\pi$ in the learned model $\hat{\mathcal{M}}$. For $d_f^\mu$, they use

$$d_f^\mu(\mathbf{s}, \mathbf{a}) = f d(\mathbf{s}, \mathbf{a}) + (1-f) d_{\hat{\mathcal{M}}}^\mu(\mathbf{s}, \mathbf{a}) \quad (38)$$

where $f \in [0, 1]$ is the ratio of data-points drawn from the offline dataset and $\mu(\mathbf{a}|\mathbf{s})$ is the rollout distribution used with the learned model. We can see $f$ as a hyperparameter that allows us to tune how conservative we want to be. Larger values of $f$ mean we will sample more from our offline dataset, and therefore, will have a more conservative $Q$ estimate in the end. Overall, these choices make it so we push down $Q$-values on state-action tuples from model rollouts and push up $Q$-values on state-action pairs from the offline dataset. Furthermore, Yu et al. [18] show that this policy evaluation step still provides a lower bound on the true $Q$-function, which is an important property to avoid OOD actions due to overestimated $Q$-values.

The main advantage of COMBO concerning MOReL and MOPO is that it removes the need for uncertainty quantification in model-based offline RL approaches, which is challenging and often unreliable. Intuitively, suppose the model produces something that looks different from real data. In that case, it is easy for the $Q$-function to make it look bad. However, if the model produces very good states and actions that are indistinguishable from the real ones, then the two regularization terms in (36) should balance out. Trabucco et al. [62] argue that this regularization term is similar to adversarial training, where we penalize OOD data with hopes of having a generator that is eventually able to fool us.

*F. One-Step Methods*

One-step methods show great promise for offline RL due to their simplicity and effectiveness. Brandfonbrener et al. [22] propose the one-step framework and experiment with multiple different policy improvement operators, ultimately showing that one-step methods can outperform multistep and iterative methods in several of OpenAI's gym environments. They attribute their success mostly to the ability to learn with very weak regularization, allowing their function approximators to fit the true $Q$-function more freely. In one of their experiments, they show how most multistep methods diverge when training with a low regularization weight, which is not able to sufficiently constrain the learned policy. Multistep methods exhibit their best performance with the smallest regularization weight that does not diverge. Increasing regularization further keeps the algorithm more stable, but results in too conservative policies.

However, Brandfonbrener et al. [22] also show that one-step methods still underperform multistep methods in some scenarios, such as when the training dataset $\mathcal{D}$ is composed largely of suboptimal behavior or when it has good coverage of the state-action space. In these cases, multistep methods do not suffer so much from iterative error exploitation and can generally fit better policies than one-step methods. Brandfonbrener et al. [22] obtain the best results when using the traditional policy evaluation objective, minimizing the Bellman error, and using exponentiated advantage estimates with maximum log-likelihood in the policy improvement step similar to (7).

Following this work, Kostrikov et al. [23] recently proposed a novel one-step method dubbed implicit $Q$-learning (IQL). Their contribution is in the policy evaluation step, where instead of updating the $Q$-function with target actions sampled from the behavior policy $\pi_\beta$, they use a function approximator for $V^\pi$ as the target, such that

$$J(\phi) = \mathbb{E}_{\mathbf{s},\mathbf{a},\mathbf{s}'\sim\mathcal{D}}\big[r(\mathbf{s},\mathbf{a}) + V_\psi^\pi(\mathbf{s}') - Q_\phi^\pi(\mathbf{s},\mathbf{a})\big]. \quad (39)$$

More critically, the state-value objective, which we also wish to minimize, can be expressed as

$$J(\psi) = \mathbb{E}_{\mathbf{s},\mathbf{a}\sim\mathcal{D}}\big[\ell\big(V_\psi^\pi(\mathbf{s}) - Q_\phi^\pi(\mathbf{s},\mathbf{a})\big)\big] \quad (40)$$

where $\ell(\cdot)$ is any error measure. If we were to use the MSE loss for $\ell$, then $V_\psi^\pi(\mathbf{s})$ will converge to $\mathbb{E}_{\mathbf{a}\sim\pi_\beta(\cdot|\mathbf{s})}[Q_\phi^\pi(\mathbf{s},\mathbf{a})]$, which satisfies the Bellman equations for action-value functions. However, ideally, we would want to satisfy the Bellman optimality equations, where $V_\psi^\pi(\mathbf{s})$ should converge to $\max_\mathbf{a} Q_\phi^\pi(\mathbf{s},\mathbf{a})$. To address this, Kostrikov et al. [23] propose using an expectile regression loss, so that we can think of $V_\psi^\pi(\mathbf{s})$ as the best value from the actions within the support of our data. The expectile loss is given by

$$\ell_2^\tau(x) = \begin{cases} (1-\tau)x^2, & \text{if } x > 0 \\ \tau x^2, & \text{otherwise} \end{cases} \quad (41)$$

where $\tau \in [0, 1]$ is a parameter we can choose to penalize negative errors much more than positive errors, making it much better for $V_\psi^\pi(\mathbf{s})$ to be larger than $Q_\phi^\pi(\mathbf{s},\mathbf{a})$, than it is to be smaller. Since this method only trains on states and actions in the dataset, we do not have to worry about overestimating $Q$-values of OOD actions. Essentially, at any given state $\mathbf{s}$, this method regresses to the best actions we have seen on similar in-distribution states, such that

$$V^\pi(\mathbf{s}) = \max_{\mathbf{a}\in\Omega(\mathbf{s})} Q^\pi(\mathbf{s},\mathbf{a}) \quad (42)$$

where $\Omega(\mathbf{s}) = \{\mathbf{a} : \pi_\beta(\mathbf{a}|\mathbf{s}) \geq \epsilon\}$. In the policy improvement step, Kostrikov et al. [23] use a procedure similar to AWR [36], with exponentiated advantage weights to extract a policy.

The key difference between IQL and the one-step methods proposed by Brandfonbrener et al. [22] is that IQL performs iterative dynamic programming. In the policy evaluation step, the value updates are based on the Bellman optimality equations, allowing us to improve the behavioral policy $\pi_\beta$. In practice, IQL has shown to be one of the most successful methods to date on the D4RL [6] benchmark, having a good and reliable performance on multiple domains with varying complexity from AntMaze to Adroit.

*G. Imitation Learning*

Most imitation learning methods are concerned with filtering out suboptimal behavior to apply a traditional supervised regression loss afterward. Chen et al. [21] propose a method named best-action imitation learning (BAIL), that fits a value function $V_\phi^\pi(\mathbf{s})$ and then uses it to select the best actions to train on. By fitting $V_\phi^\pi(\mathbf{s})$ to the approximate upper envelope of dataset $\mathcal{D}$, one can approximate the optimal value function that satisfies the Bellman optimality equations. To this end, Chen et al. [21] propose to minimize the objective

$$J(\phi) = \mathbb{E}_{\tau\sim p_{\pi_\beta}(\cdot)}\left[\sum_{t=0}^H \big(V_\phi^\pi(\mathbf{s}_t) - R_t\big)^2 w(\mathbf{s}_t)\right] \quad (43)$$

where

$$w(\mathbf{s}_t) = \begin{cases} 1, & \text{if } V_\phi^\pi(\mathbf{s}_t) < R_t \\ K, & \text{otherwise.} \end{cases} \quad (44)$$

Intuitively, when we set $K \gg 1$, this objective will penalize the value function much more heavily when it is further away from good return samples $R_t$, approximating the upper envelope of returns in $\mathcal{D}$. After fitting $V_\phi^\pi(\mathbf{s})$, BAIL selects the best state-action pairs $(\mathbf{s}, \mathbf{a})$ from $\mathcal{D}$ with returns greater than a chosen ratio of the estimated value-function, that is,

$$R_t > \rho V_\phi^\pi(\mathbf{s}_t) \quad (45)$$

where they use $\rho = 0.25$. Finally, they use the filtered state-action pairs to learn a policy through BC.

Siegel et al. [46] propose a method that learns an Advantage-weighted Behavior Model (ABM) and uses it prior to performing maximum a posteriori policy optimization [63] (MPO). Their algorithm consists of multiple iterations of policy evaluation and prior learning until they finally perform a policy improvement step with their learned prior to extracting the best possible policy. The policy evaluation step fits a $Q$-function using standard bootstrapped regression seen in (18). The prior $\pi_{\theta_{\text{abm}}}$ is learned using an $n$-step advantage estimator $A_n^\pi$, such that

$$A_n^\pi(\mathbf{s}_t, \mathbf{a}_t) = \sum_{t'=t}^{t+n} \gamma^{t'-t} r(\mathbf{s}_{t'}, \mathbf{a}_{t'}) + \gamma^n V_\phi^\pi(\mathbf{s}_{t+n}) - V_\phi^\pi(\mathbf{s}_t) \quad (46)$$

where $V_\phi^\pi(\mathbf{s}_t)$ is approximated as the sampled mean of the learned action-value function $Q_\phi^\pi(\mathbf{s}_t, \mathbf{a}_t)$. The advantage estimator is used to filter out the suboptimal trajectories in a manner akin to implicit policy constraints from (7), where we use an indicator function instead of an exponential, such that

$$J(\theta_{\text{abm}}) = \mathbb{E}_{\mathbf{s},\mathbf{a} \sim \mathcal{D}} \left[ \log \pi_{\theta_{\text{abm}}}(\mathbf{a}|\mathbf{s}) \mathbb{1}\left[\hat{A}^\pi(\mathbf{s}, \mathbf{a}) > 0\right] \right]. \quad (47)$$

After learning the ABM prior, Siegel et al. [46] use MPO to learn an improved policy $\pi_\theta$, constrained to $\pi_{\theta_{\text{abm}}}$ through a KL divergence. They show that this approach is able to outperform methods that try to directly learn $\pi_\theta$ using a KL constraint with $\pi_\beta$ since this penalizes the model when $\pi_\beta$ is comprised of suboptimal behavior.

Furthermore, Wang et al. [45] propose another method named critic regularized regression (CRR), which also uses the indicator function to aggressively filter out below-average actions like in (47), but opts for a more pessimistic advantage estimator

$$A_{\text{CRR}}^\pi(\mathbf{s}, \mathbf{a}) = Q_\phi^\pi(\mathbf{s}, \mathbf{a}) - \max_{\tilde{\mathbf{a}}} Q_\phi^\pi(\mathbf{s}, \tilde{\mathbf{a}}) \quad (48)$$

where we change the expectation to a max operator in the AWAC advantage estimator from (19). According to Wang et al. [45], this approach seems to outperform implicit policy constraint methods on tasks that have a mix of expert and suboptimal behavior, since policy constraint methods tend to be too permissive and copy inferior actions as the policy improves.

Toward learning conditional policies, Emmons et al. [47] proposed RL via supervised learning (RvS), which uses the common framework for conditional BC methods from (14). They propose learning policies with two different types of outcomes, a goal-conditioned policy and a reward-conditioned one. The outcomes for the goal-conditioned policy are sampled from

$$g_g(\omega|\tau_{t:H}) = \mathcal{U}(\mathbf{s}_{t+1}, \mathbf{s}_H) \quad (49)$$

while the reward-conditioned policy's outcomes are sampled from

$$g_r(\omega|\tau_{t:H}) = \mathbb{1}\left[\omega = \frac{1}{H-t+1} \sum_{t'=t}^{H} r(\mathbf{s}_{t'}, \mathbf{a}_{t'})\right]. \quad (50)$$

Surprisingly, Emmons et al. [47] show that this simple formulation combined with an increased network capacity and regularization is capable of learning policies that are as good or better than more complex approaches (e.g., CQL) in some domains. However, the choice of conditioning variable is crucial for the method's performance. Conditioning on goals, as in $g_g$, performs well in environments that benefit from compositionality, such as AntMaze and FrankaKitchen. In contrast, conditioning on rewards, as in $g_r$, performs very poorly in these domains but attains a good performance in the Gym-MuJoCo suite, where $g_g$ is not applicable.

### H. Trajectory Optimization

Janner et al. [28] recently proposed a trajectory optimization method they call trajectory transformer (TT), that uses a transformer architecture [37] to model the trajectory distributions $p_{\pi_\beta}(\tau)$. In their formulation, they represent a trajectory as a sequence of states and actions interleaved by returns-to-go, that is,

$$\tau = (R_0, \mathbf{s}_0, \mathbf{a}_0, R_1, \mathbf{s}_1, \mathbf{a}_1, \ldots, R_H, \mathbf{s}_H). \quad (51)$$

During training, they sample trajectories from $\mathcal{D}$ and maximize the log-likelihood of each token from the sequence (e.g., $R_0$, $\mathbf{s}_0$, $\mathbf{a}_0$, and so on) conditioned on all previous ones. Once they learn a model of the trajectory distribution $p_{\psi_\tau}$, they use beam search [64] together with the reward-to-go estimates $\hat{R}_t$ for planning.

Concurrently, Chen et al. [27] proposed a trajectory optimization method for offline RL named decision transformer (DT), which is also based on transformers. Their formulation uses a similar trajectory representation as Janner et al.'s [28] in (51), but has a modified training procedure and opts for a different planning strategy. During training, instead of increasing the maximum log-likelihood of all tokens, they focus exclusively on minimizing the MSE between the predicted and ground-truth actions from the trajectory, arguing that learning to predict states and returns-to-go are not necessary for good performance. During the evaluation, instead of using beam search for planning, Chen et al. [27] condition the trajectory rollouts on a specified target return based on the desired performance on a given task (e.g., the maximum possible return to generate expert behavior). The DT then selects an initial action based on the target return and initial state, observes the new state and reward, and updates its reward-to-go target by subtracting the observed reward. Although expensive to train, these methods perform well in sparse reward settings, where temporal-difference methods typically fail, since they rely on dense reward estimates to effectively propagate $Q$-values over long horizons.

## V. Off-Policy Evaluation

One of the biggest open problems in offline RL is hyperparameter tuning [8]. Determining the set of hyperparameters that yield the best possible policy in an offline manner is extremely valuable, allowing us to save valuable resources and avoid dangerous interactions from online interaction. OPE is the task of evaluating a policy only through previous experiences. Excessively training on the same offline dataset can lead to poor solutions due to overfitting. Hence, it is paramount to find good OPE methods that allow us to validate our policy during training [65]. However, in practice, most offline RL methods do not rely on OPE methods to evaluate performance and instead train with a set of hyperparameters for a fixed number of steps and use the policy from the last iteration to evaluate *online* their quality. We consider the development of OPE methods essential for the development of offline RL and dedicate a part of this survey to discuss its developments.

We detail three of the most popular OPE methods in Sections V-A–V-C. Let $\hat{J}(\pi)$ denote the OPE objective used to evaluate a policy $\pi$ and $\mathcal{D}_e$ the static evaluation dataset. We have the following OPE methods:

### A. Model-Based Approach

In the model-based approach, we first fit the model dynamics $T_{\psi_T}(\mathbf{s}_{t+1}|\mathbf{s}_t, \mathbf{a}_t)$ and reward function $r_{\psi_r}(\mathbf{s}_t, \mathbf{a}_t)$ using $\mathcal{D}_e$, where $\psi_T$ and $\psi_r$ are the parameters of the learned dynamics and rewards model, respectively. Let $p_{\psi_T}(\tau)$ denote the trajectory distribution induced by following policy $\pi$ with transition dynamics $T_{\psi_T}$. We can evaluate the policy by computing the expected return under $p_{\psi_T}(\tau)$, such that

$$\hat{J}(\pi) = \mathbb{E}_{\tau \sim p_{\psi_T}(\cdot)} \left[ \sum_{t=0}^{H} \gamma^t r_{\psi_r}(\mathbf{s}_t, \mathbf{a}_t) \right]. \tag{52}$$

### B. Importance Sampling

With IS, we first fit an estimate of the behavior policy $\hat{\pi}_\beta(\mathbf{a}|\mathbf{s})$ using $\mathcal{D}_e$. Then, we compute the expected return under our policy $\pi$ by evaluating the importance-sampling objective

$$\hat{J}(\pi) = \mathbb{E}_{\tau \sim p_{\hat{\pi}_\beta}(\cdot)} \left[ w_{0:H} \sum_{t=0}^{H} \gamma^t r(\mathbf{s}_t, \mathbf{a}_t) \right] \tag{53}$$

where $w_{i:j} = (\prod_{t=i}^{j} \pi(\mathbf{a}_t|\mathbf{s}_t))/(\prod_{t=i}^{j} \hat{\pi}_\beta(\mathbf{a}_t|\mathbf{s}_t))$ is the product of the importance weights. Here, we can also use any of the variance-reduction strategies for IS reviewed in Section IV-B (e.g., weighted [40], doubly robust [41], or marginalized IS [24], [25], [26]).

### C. Fit Q Evaluation

In fit $Q$ evaluation (FQE), we first train a $Q$-function $Q_\phi^\pi$ by minimizing the Bellman error under the policy $\pi$. Then, we evaluate the policy by computing the average expected return over the states and actions from $\mathcal{D}_e$, such that

$$\hat{J}(\pi) = \mathbb{E}_{\mathbf{s},\mathbf{a} \sim \mathcal{D}_e} \left[ Q_\phi^\pi(\mathbf{s}, \mathbf{a}) \right]. \tag{54}$$

Toward determining the best OPE approach, Voloshin et al. [66] present a comprehensive empirical study of several different methods. They evaluate 33 different OPE methods using a relative MSE metric between the estimated on-policy value and the true on-policy value. Their study shows that FQE performs surprisingly well, despite its simplicity.

In a different study, Paine et al. [65] review the effectiveness of different OPE methods for hyperparameter selection. They evaluate different strategies on complex environments available in the RL Unplugged [35] benchmark. The work shows that using policy constraint algorithms, like CRR [45], reestimating the $Q$ values using FQE, and using $\hat{V}(\mathbf{s}_0)$ as a ranking statistic is sufficient for performing offline hyperparameter selection. More recently, Fu et al. [67] conducted a new study of OPE strategies and proposed the novel deep OPE (DOPE) benchmark to help accelerate the development of OPE methods. Despite all methods achieving suboptimal performance, FQE seems to have the best overall performance on benchmarks like RL Unplugged [35] and D4RL [6].

Ultimately, in real-world settings, we often cannot roll out our policy to evaluate if it will work or not. This is a considerable barrier to the practical use of offline RL methods since we still rely too heavily on simulator rollouts to verify that a method works. Therefore, having robust OPE methods that work reliably across a wide variety of datasets is essential for advancing the field.

## VI. Benchmark Review

In Sections VI-A and VI-B, we review the two most widely accepted benchmarks for offline RL and the single benchmark for OPE, respectively, discussing their properties and limitations. In Section VI-C, we also cover the performance of current methods on the D4RL benchmark to give readers a better sense of which methods have shown the best performance to date.

### A. Offline RL Benchmarks

Prior work on offline RL [15], [16], [36], [57] has typically used an online RL algorithm to train the behavior policy $\pi_\beta$ and opted to either use data from the replay buffer or rollouts from the final policy to create the static dataset $\mathcal{D}$. In practice, data might come from non-Markovian policies, such as human agents, or hand-engineered policies, making datasets based on online RL algorithms unrepresentative of the situations we might have to deal with in the real world. Ideally, we would use real-world datasets to evaluate our offline RL algorithms. However, evaluating a candidate policy is difficult since we might have to take actions outside of the support of our dataset, which can be dangerous in areas like autonomous driving and medical diagnosis. Although one could use OPE methods outlined in Section V, these methods are still too unreliable for one to make confident predictions.

In Section VI-A1, we outline the properties offline datasets must have to provide a meaningful measure toward progress in realistic applications of offline RL. Section VI-A2 presents the two largest offline RL benchmarks to date: D4RL [6] and RL Unplugged [35], with an overview of the environments in each benchmark and the properties that they satisfy. Finally, in Section VI-A3, we summarize some of the missing properties of the current offline RL benchmarks.

*1) Dataset Design Factors:* In this subsection, we outline the desired properties for offline datasets, according to Fu et al. [6] and Gulcehre et al. [35], to provide a meaningful measure toward progress in realistic applications of offline RL. These properties include:

1) *Narrow and Biased Data Distributions (NB):* This can arise in human demonstrations or when using hand-crafted policies. It is important for offline RL not to diverge in these cases and avoid visiting too many OOD states.
2) *Undirected and Multitask Data (UM):* This is important to assess the algorithm's ability to perform stitching, i.e., combining portions of existing trajectories in order to solve a task even if none of the individual trajectories are solutions to the task at hand. This property can naturally arise when data is passively logged or when we want to propose goals to an agent different from the ones used to collect the trajectories. Algorithms that do not perform multistep dynamic programming and are based on constrained or regularized approximation have a particularly hard time recovering the optimal policy from undirected data.
3) *Sparse Rewards (SR):* This can be challenging due to the difficulty of credit assignment. Manually engineering a reward function that aligns with the task is often tricky and can lead to solutions that exploit local optima. Designing sparse rewards is typically easier since it only requires defining the criteria for solving a task, making it an appealing property to address.
4) *Suboptimal Data (SD):* This is important to assess an algorithm's ability to generalize beyond imitation learning. Suboptimal data leaves room for improvement in the learned policy and allows us to evaluate an algorithm's ability to generalize and filter out bad behaviors from the dataset. Models typically have a hard time with generalization, which makes it difficult to improve beyond the underlying suboptimal behavior policy.
5) *Nonrepresentable Behavior Policies (NR):* This arises when the function approximator cannot fully capture the underlying behavior's complexity. Ultimately, we must work with a projection of the optimal policy to our policy space and must handle scenarios where our policy space cannot represent the true policy. RL implementations typically default to networks with a few dense layers to represent their policy. Experimenting with different network architectures can be critical in datasets with nonrepresentable behavior policies to obtain the best possible projection.
6) *Non-Markovian Behavior Policies (NM):* This naturally arises in behavior policies from human agents or hand-engineered controllers. Offline RL algorithms should be susceptible to violations of the Markovian property if we expect to apply them to real-world datasets from human agents in the future. One of the challenges with these datasets is figuring out what state representation to use that best approximates the Markov property.
7) *Realistic Domains (RD):* Using real-world environments is infeasible for an RL benchmark since results would be too hard to reproduce and likely inaccessible to most of the public. It is important to have simulated environments with high fidelity to real-world behaviors to ensure that offline RL algorithms address issues that come up in deployment. Models usually have difficulty learning optimal policies when subject to noisy readings or imperfect actuators, which can often occur in real scenarios.
8) *Nonstationarity (NS):* An agent may experience settings where sensors malfunction, actuators degrade or reward functions are updated, causing perturbations in the MDP that vary over time (e.g., as a pump's efficiency degrades over time). To account for this, models need a strategy to select sub-policies and apply them in the correct time step.

Besides these desirable properties, we also have environment characteristics that we wish to find in the datasets, including continuous action and state spaces, stochastic dynamics, and partial observability. Continuous spaces are often more challenging than discrete ones since it is infeasible to visit every state in a continuous domain, forcing an agent to generalize beyond seen states and actions. Although pixel observations are technically discrete, they are often considered just as or even more challenging than continuous observations due to the many dimensions. Stochastic dynamics are also desirable since they are more common in the natural world, where there is an inherent randomness to events normally due to model limitations. Partial observability often arises when we lack domain knowledge to observe the true state of an environment. Ensuring offline RL works under POMDPs is essential for its application in the real world.

*2) Datasets Overview:* Between the D4RL [6] and RL Unplugged [35] benchmarks, we have environments and datasets that tackle most of the properties that characterize a good offline RL dataset. The D4RL benchmark includes datasets for OpenAI's Gym-MuJoCo tasks, mazes, dexterous manipulation tasks (i.e., Adroit [68]), robotic manipulation tasks (i.e., FrankaKitchen [69]), autonomous driving (i.e., car learning to act (CARLA) [70]), and traffic simulation (i.e., flow [71]). The RL Unplugged benchmark consists of datasets from four different suites, including the DeepMind control Suite [72], DeepMind locomotion [73], arcade learning environment [74] (ALE), and real-world RL suite [75]. In Table III, we characterize all of the environments available in D4RL and a few key ones from RL Unplugged, showing the types of spaces, dynamics, MDPs, and properties from Section VI-A1 that each environment's datasets satisfy.

Although RL Unplugged has a wide variety of tasks, one fundamental issue with the benchmark is that all behavior policies come from actors trained online. While D4RL has policies that are nonrepresentable by design (e.g., non-Markovian), RL Unplugged has no such guarantee, making it likely that the behavior policy is often representable due to the use of similar network architectures in the field. Furthermore, since the trajectories in the dataset are randomly sampled from the replay buffer of an agent trained online, the datasets will typically have trajectories that solve the task at hand and there are no guarantees that the algorithm must perform stitching to succeed in the environment.

Another difference between both benchmarks lies in their evaluation protocols. While D4RL does not impose a particular evaluation protocol, RL Unplugged separates the benchmark environments into online and offline validation environments. The online environments allow algorithms to use online samples for validation, which does not fully capture the essence of offline RL since one of the premises is that online interactions are likely to be costly. Evaluating an algorithm trained with online validation still makes sense in scenarios in which online validations are not prohibitively expensive, and we still want to leverage a large static dataset of previously collected data for training. The offline validation environments require one to use OPE strategies to evaluate the performance of their method and perform hyperparameter tuning.

Additionally, D4RL makes sure to provide datasets collected with random, medium, and expert policies in some environments, allowing us to evaluate whether an algorithm can extract meaning from noise. On the other hand, RL Unplugged mostly limits its datasets to behavior policies that have been successfully trained with an online agent, such that most of the data comes from medium to expert policies.

*3) Missing Properties:* Current benchmarks still have insufficient datasets with stochastic dynamics in the environment (except for the Atari suite), common in real-world settings (e.g., economics, healthcare, education, and so on) and essential to evaluate. Environments that are nonstationary (i.e., change over time) are also very common in the real world and still have limited coverage in current benchmarks, only being present in the real-world RL suite. Furthermore, datasets designed to have risky biases are also important. For instance, if your driving data never shows a car crash, an offline RL algorithm should still be able to learn how to avoid car crashes. Finally, we have not found any offline RL datasets for multiagent environments, which may arise in settings like robot team navigation, smart grid operation, and control of mobile sensor networks [76].

*B. Off-Policy Evaluation Benchmarks*

OPE is the problem of evaluating the expected performance of a method using only offline data. This is important for several reasons, including providing high-confidence guarantees before deployment [77], performing hyperparameter tuning [65], and determining when to stop training a given model to avoid overfitting [78]. The DOPE [67] benchmark was created to provide a standardized framework for comparing OPE algorithms by providing tasks with a wide range of difficulty that satisfy desirable design properties and a set of policies with different behavioral patterns. In Section VI-B1, we provide a brief overview of the DOPE benchmark. Then, in Section VI-B2, we cover the evaluation metrics used to compare different OPE methods.

*1) Benchmark Overview:* The DOPE benchmark is divided into two domains: DOPE RL unplugged and DOPE D4RL, each with its own set of datasets and policies. In the DOPE RL unplugged domain, the datasets are generated from the experiences collected from an online RL algorithm, as we explained in Section VI-A2. The policies are generated from offline RL algorithms trained on these datasets. The algorithms are chosen to ensure that evaluation policies differ from the behavior policies, and multiple policy snapshots are saved at exponentially increasing intervals. In the DOPE D4RL domain, the datasets are built from a mixture of random exploration policies, human demonstrations, non-Markovian controllers, and online RL algorithms, making it more reflective of practical settings. Furthermore, the policies are generated using online RL algorithms, making it less likely that the evaluation policies will have similar state-action distributions to the behavior policies and exacerbate distributional shifts. This allows the DOPE benchmark to cover both idealized and practical data settings with a wide range of difficulty for both.

The benchmark also provides six baseline OPE methods for comparison, three of which we already discussed in Section V: fit $Q$-evaluation (FQE), model-based (MB), IS), doubly robust (DR), DICE, and variational power method (VPM). Across all metrics and most datasets, MB and FQE have performed the best. However, no method seems to perform consistently better in all settings.

*2) Evaluation Metrics:* In OPE, we can have different objectives we wish to meet. One is to estimate the performance, or value, of a policy $\pi$, such that the estimated value is as close as possible to the true value $V^\pi$ of our policy. Another objective is to select the best possible policy among a set of candidate policies. In this case, we are only interested in estimating the relative value between policies instead of their absolute value. This second objective is useful in hyperparameter tuning and early stopping during training. However, when deploying our policy to the real world, we might still need an absolute measure of its quality to assess the cost and danger of such deployment. Here, we list the three evaluation metrics from the DOPE benchmark that allow one to perform OPE and selection.

*a) Absolute error:* This metric is intended for OPE instead of selection. Fu et al. [67] opt to use the absolute error instead of the MSE to increase robustness to outliers.

*b) Regret@k:* This metric is intended for off-policy selection. It evaluates the difference in value between the best policy among the estimated best $k$ policies and the actual best policy in the set.

*c) Rank Correlation:* This metric is intended for off-policy selection. It computes the correlation between the ordinal rankings according to OPE estimates and the true ordinal rankings of the policies.

*C. Method Performance*

To determine which methods are the most promising, we wish to evaluate their performance on the benchmarks from Section VI-A. Since most of the works that we found do not provide results for the datasets in RL Unplugged [35], we only used the scores found for D4RL [6]. Fig. 5 provides the relative scores of various methods and taxonomy classes on each of the dataset properties from Section VI-A1. From the heatmap, we can see that several methods do a poor job of evaluating a variety of datasets, which hinders our ability to compare their performance. The lack of datasets with nonstationarity data in D4RL is also harmful to the field since most methods do not bother evaluating datasets outside D4RL.

In general, given that the methods are ordered from left to right by release date, we can observe from the leftmost heatmap that recent methods tend to outperform older ones across all datasets. Methods like TT [28] and implicit $Q$-learning [23] are currently among the best performing.

The rightmost heatmap shows the relative performance of different taxonomy classes on each dataset property, where the scores were aggregated by computing the max score of any given method that belongs to such class. From the figure, we can see that the best-performing classes are trajectory

TABLE III
DIFFERENT RL ENVIRONMENTS USED TO CREATE OFFLINE RL DATASETS AVAILABLE IN THE D4RL [6] AND RL UNPLUGGED [35] BENCHMARKS. WE CATEGORIZE EACH ENVIRONMENT WITHIN ITS TASK SUITE AND BENCHMARK, WHEN APPLICABLE. FOR EACH ENVIRONMENT, WE SPECIFY WHETHER THE ACTION AND STATE SPACE ARE CONTINUOUS OR DISCRETE, WHETHER THE DYNAMICS ARE DETERMINISTIC OR STOCHASTIC, WHETHER THE MDP IS FULLY OR PARTIALLY OBSERVABLE, AND EACH OF THE SATISFIED DESIRABLE PROPERTIES OF OFFLINE RL DATASETS FROM SECTION VI-A1. FOR THE SUITES THAT HAVE TOO MANY ENVIRONMENTS (E.G., ADROIT), WE ONLY LIST THREE FROM EACH FOR BREVITY. FOR THE FULL LIST OF ENVIRONMENTS IN RL UNPLUGGED, REFER TO GULCEHRE ET AL.'S [35] PAPER

| Benchmark | Suite | Environment Name | $\mathcal{A}$ | $\mathcal{S}$ | $T(\mathbf{s'}|\mathbf{s},\mathbf{a})$ | $\mathcal{M}$ | NB | UM | SR | SD | NR | NM | RD | NS |
|---|---|---|---|---|---|---|---|---|---|---|---|---|---|---|
| D4RL | | Maze2D | cont | cont | det | full | | ✓ | | | | ✓ | | |
| | | AntMaze | cont | cont | det | full | | ✓ | ✓ | | | ✓ | | |
| | Gym-MuJoCo | HalfCheetah | cont | cont | det | full | ✓ | | | ✓ | | | | |
| | | Hopper | cont | cont | det | full | ✓ | | | ✓ | | | | |
| | | Walker2d | cont | cont | det | full | ✓ | | | ✓ | | | | |
| | Adroit | Pen | cont | cont | det | full | ✓ | | ✓ | | ✓ | | ✓ | |
| | | Hammer | cont | cont | det | full | ✓ | | ✓ | | ✓ | | ✓ | |
| | | Door | cont | cont | det | full | ✓ | | ✓ | | ✓ | | ✓ | |
| | | FrankaKitchen | cont | cont | det | full | ✓ | | | | | | ✓ | |
| | | Flow | cont | cont | det | full | | | | | ✓ | | ✓ | |
| | | CARLA | cont | pixel | det | partial | ✓ | | | | ✓ | | ✓ | |
| RL Unplugged | DM Control | Finger turn hard | cont | cont | det | full | | | | | | | | |
| | | Fish swim | cont | cont | det | full | | | ✓ | | | | | |
| | | Manipulator insert peg | cont | cont | det/stoch | full | | | ✓ | ✓ | | | ✓ | |
| | DM Locomotion | Humanoid gaps | cont | pixel | det | full | | | | | | | ✓ | |
| | | Rodent bowl escape | cont | pixel | det | partial | | | | | | | | |
| | | Rodent mazes | cont | pixel | det | partial | | | | | | | | |
| | Atari | Space Invaders | disc | pixel | stoch | full | | | | | | | | |
| | | Pong | disc | pixel | stoch | full | | | ✓ | | | | | |
| | | Breakout | disc | pixel | stoch | full | | | | | | | | |
| | Real world | Cartpole swingup | cont | cont | det | partial | | | ✓ | ✓ | | | ✓ | ✓ |
| | | Walker walk | cont | cont | det | partial | | | ✓ | ✓ | | | ✓ | ✓ |
| | | Humanoid walk | cont | cont | det | partial | | | ✓ | ✓ | | | ✓ | ✓ |

∿ Continuous space   ⊓ Discrete space   🖼 Pixel space   ⊡ Stochastic dynamics   ◉ Deterministic dynamics   👁 Fully observable   📷 Partially observable optimization and one-step methods combined with implicit policy constraint and value regularization elements. Although most methods and classes tend to perform similarly across all types of datasets, we found that trajectory optimization methods perform particularly well in scenarios with sparse rewards and undirected and multitask data. This shows that planning can be compelling in offline RL, especially when combined with $Q$-functions trained via dynamic programming. This is the case for TT [28], which uses a $Q$-function trained with IQL [23] to guide the planning procedure.

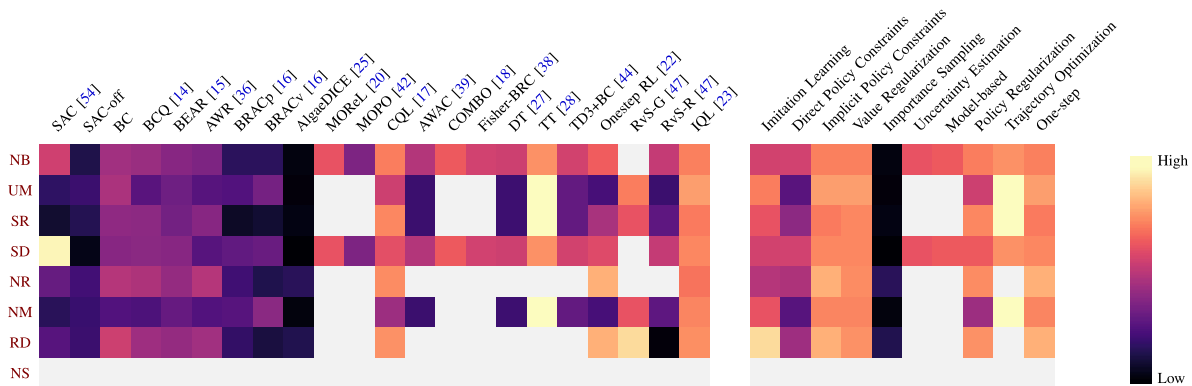

Fig. 5. Relative performance of different offline RL methods (left) and offline RL taxonomy classes (right) on the dataset properties outlined in Section VI-A1. Brighter colors signify a higher performance. Light gray indicates the method was not evaluated on any datasets that satisfy the given property. On the right, the taxonomy classes show the aggregated performance of each classification from Table II. The class score on a given dataset property is given by the maximum score of any algorithm belonging to that class on the same property. The relative score for a dataset property is computed based on the average normalized D4RL scores on all datasets that satisfy such property. The normalized D4RL score is on a relative scale $\in [0, 1]$, where these are lowest and highest scores among all algorithms evaluated on a given dataset. Refer to the Supplementary Materials (🔗) for more details on the raw scores used to generate this figure.

## VII. PRACTICAL APPLICATIONS

Most of the breakthroughs in online RL were developed in simulation, where data is effectively unlimited, and there are no consequences for poor actions. Applying these algorithms to the real world has proven incredibly challenging since many interesting systems are typically too complex to simulate [79]. One of the appeals of offline RL is its ability to learn a policy using previously collected data without the risk or expense of interacting with the real world. Levine et al. [8] and Fu and Di [80] have extensively covered multiple real-world applications of offline RL, including robotics [9], autonomous driving [12], [81], healthcare [11], [82], dialog systems [83], and energy management systems [84].

Here, we highlight, through recent examples, a few reasons one might use offline RL over online RL in a given application. In healthcare, Emerson et al. [82] used offline RL to develop a policy that selects the optimal insulin dose to maintain blood glucose levels within a healthy range. They argue that online RL is far too unstable to manage glucose levels and could cause patients to go outside of their healthy threshold. In energy management, Zhan et al. [84] propose a model-based offline RL algorithm to optimize the combustion control strategy for thermal power generating units (TPGUs). By combining large amounts of historical data from TPGUs and low-fidelity simulation data, they can learn a safety-constrained policy that far surpasses BC. In this case, it was far less expensive and time-consuming to leverage the existing data to learn a policy instead of doing so interactively. Finally, Verma et al. [83] propose using offline RL to train a task-driven dialog agent named CHAI (CHatbot AI). Applying online RL to dialog systems can be prohibitively expensive due to the cost of interacting with a human, and using simulated human agents typically requires strong priors to work. CHAI leverages the vast amounts of unlabeled dialog data and labeled task-driven data to learn a dialog agent that is more effective than those previously trained with online RL.

## VIII. Open Problems

Several of the open problems of the offline RL field listed in Levine et al.'s [8] work remain to this date. However, some of these problems have seen considerable progress. This section provides an update on the open problems and future directions of the field.

Hyperparameter tuning [65] and OPE are two problems that still lack a satisfying solution. Currently, we either use inaccurate OPE methods for hyperparameter tuning or train for a fixed number of steps. These are both lackluster approaches since we are often left with a suboptimal policy that might have to overfit our data. Finding good ways to validate policies will also benefit training, allowing us to early stop training that exhibits degrading performance over time. Levine et al. [8] argued that shifting toward off-policy selection instead of the evaluation was a promising direction for OPE methods. While we have seen this shift occur with the introduction of the DOPE benchmark, we still lack a method that can consistently outperform the others on most datasets.

Emerging areas in RL, like incremental RL [85], are being developed in parallel and are promising for offline RL's future development. Incremental RL directly contributes to solving offline RL problems with nonstationary datasets and developing online fine-tuning strategies that use offline policies.

Safety-critical RL is also an area we wish to see more people tackle and benchmark in the future. Strategies like uncertainty estimation and regularization have been used to avoid OOD states, but a few works take into account avoiding safety-critical in-distribution states. Toward this end, some works [86] use a conditional value-at-risk (CVaR [87]) objective to learn a risk-averse policy.

Finally, a promising future direction for the field is the use of unsupervised RL techniques to leverage large amounts of unlabeled data. In many cases, labeling large datasets with rewards may be costly, especially if these require human supervision [88]. Leveraging diverse unlabeled data in a simple yet effective manner is still an open problem. Yu et al. [89] show how it is possible to learn effective policies from large amounts of suboptimal unlabeled data combined with a limited amount of high-quality labeled data. Kumar et al. [90] present a similar result when comparing offline RL to BC methods. More surprisingly, Yarats et al. [91] show how one can use diverse unlabeled datasets with downstream reward relabeling to achieve better performance with vanilla off-policy RL methods [51] in offline settings. Although a lot of focus has been placed on the development of new algorithms, these works show how the data we use to train these algorithms can be just as important for their performance. Finding new exploratory techniques to collect the data and novel ways to leverage unlabeled data can help extend offline RL's applicability to even more real-world domains.

## IX. Conclusion

In this survey, we provide a comprehensive overview of offline RL. First, we present a novel taxonomy to classify all offline RL methods and a set of optional modifications that can be made to each class. We also review the main offline RL methods from each class in our taxonomy and the main benchmarks in the field, including offline RL and OPE benchmarks. Finally, we share our perspective on the open problems of the field, including promising future directions for research.

## Acknowledgment

The project was coordinated by Softex, Austin, TX, USA, and published as intelligent agents for mobile platforms based on cognitive architecture technology [01245.013778/2020-21].

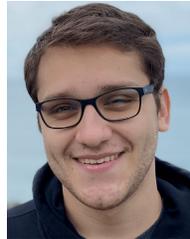

**Rafael Figueiredo Prudencio** received the B.Sc. degree (summa cum laude) in computer engineering from the University of Campinas, Campinas, Brazil, in 2021, where he is currently pursuing the M.Sc. degree in computer science.

He is currently a Software Engineer with Meta Platforms Technologies U.K., Ltd., London, U.K. His research interests include eye tracking, visual odometry, autonomous robotics, and reinforcement learning.

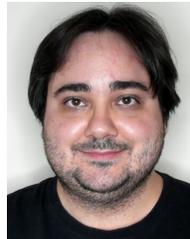

**Marcos R. O. A. Maximo** received the B.Sc. degree (summa cum laude) in computer engineering and the M.Sc. and Ph.D. degrees in electronic and computer engineering from the Aeronautics Institute of Technology (ITA), São José dos Campos, Brazil, in 2012, 2015, and 2017, respectively.

He is currently a Professor with ITA, where he is also a member of the Autonomous Computational Systems Laboratory and the Leader of the ITAndroids robotics competition team. His research interests include humanoid robotics, mobile robotics, dynamic systems control, and artificial intelligence.

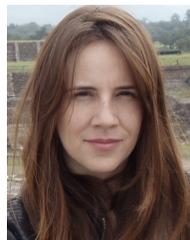

**Esther Luna Colombini** (Member, IEEE) received the M.Sc. and Ph.D. degrees in computing engineering from the Aeronautics Institute of Technology, São José dos Campos, Brazil, in 2005 and 2014, respectively.

She is currently a Professor with the University of Campinas, Campinas, Brazil, where she also coordinates the Laboratory of Robotics and Cognitive Systems. She is also a Research Member with the Advanced Institute for Artificial Intelligence, São Paulo, Brazil, the Brazilian Institute of Neuroscience and Neurotechnology, São Paulo, and the Center for Artificial Intelligence, São Paulo. She acted as the President of RoboCup Brazil, and a Co-Founder and the Chair of the Brazilian Robotics Olympiad and the National Robotics Fair, two public initiatives—supported by the National Council of Technological and Scientific Development—reaching yearly over 230 000 students aiming at attracting them to STEM careers. She coordinates the multilateral BRICSmart Alliance supported by CNPq and the BRICS-STI committee to employ ICT for smart resource utilization to combat global pandemic outbreaks and the learning in cognitive architectures research in the hub for artificial intelligence and cognitive architectures. Her research interests include machine learning and AI for autonomous robotics and healthcare, focusing on reinforcement learning, attentive models, and cognitive modeling.